\documentclass{article}

    \PassOptionsToPackage{numbers, compress}{natbib}

\usepackage[eandd, final]{neurips_2026}

\usepackage[utf8]{inputenc} 
\usepackage[T1]{fontenc}    
\usepackage[pagebackref=false,breaklinks=true,colorlinks,bookmarks=false]{hyperref}
\hypersetup{linkcolor=[rgb]{0.7,0.1,0.1}}
\hypersetup{citecolor=[rgb]{0.4,0.15,0.95}}

\usepackage{xcolor}         

\let\xfootnote\footnote        

\newcommand\blfootnote[1]{%
  \begingroup
  \renewcommand\thefootnote{}\xfootnote{#1}%
  \addtocounter{footnote}{-1}%
  \endgroup
}

\renewcommand{\footnote}[2][]{} 

\title{RxEval: A Prescription-Level Benchmark for Evaluating LLM Medication Recommendation}

\usepackage{xspace}
\newcommand{\methodName}{\mbox{RxEval}\xspace}
\author{%
  Shuhao~Chen\textsuperscript{1,2},
  Weisen~Jiang\textsuperscript{3},
  Changmiao~Wang\textsuperscript{4},
  Xiaoqing~Wu\textsuperscript{5},
  Xuanren~Shi\textsuperscript{5}, \\
  \textbf{Yu~Zhang\textsuperscript{2,\,$\dagger$},
  James~T.~Kwok\textsuperscript{1}} \\[4pt]
  $^1$The Hong Kong University of Science and Technology \\
  $^2$Southern University of Science and Technology \\
  $^3$The Chinese University of Hong Kong \\
  $^4$The Chinese University of Hong Kong, Shenzhen \\
  $^5$Shenzhen University General Hospital \\
  {\tt\small shuhochen@gmail.com}
}

\usepackage{microtype}
\usepackage{graphicx}
\usepackage{subcaption}
\usepackage{caption}
\usepackage{booktabs} 
\usepackage[pagebackref=false,breaklinks=true,colorlinks,bookmarks=false]{hyperref}
\hypersetup{linkcolor=[rgb]{0.7,0.1,0.1}}
\hypersetup{citecolor=[rgb]{0.4,0.15,0.95}}

\usepackage{algorithm}
\usepackage{algorithmic}

\usepackage{colortbl}

\usepackage{amsmath}
\usepackage{amssymb}
\usepackage{mathtools}
\usepackage{amsthm}

\usepackage[capitalize,noabbrev]{cleveref}
\usepackage{colortbl}
\usepackage{nicematrix}
\usepackage[inline, shortlabels]{enumitem}
\usepackage{multirow}

\usepackage{amsthm}
\usepackage{amsmath,amsfonts,bm}
\usepackage{nicefrac}
\usepackage{array}
\usepackage{wrapfig}

\usepackage{url}
\usepackage{pbox}

\definecolor{bluex}{rgb}{0.27, 0.42, 0.81}
\definecolor{purplex}{HTML}{9564bf}
\definecolor{red3}{HTML}{C52A20}
\definecolor{red2}{HTML}{B36A6F}
\definecolor{red1}{HTML}{FFb5b5}
\definecolor{purple}{HTML}{B36A6F}
\definecolor{darkyellow}{HTML}{D5BA82}
\definecolor{blue1}{HTML}{508AB2}
\definecolor{blue2}{HTML}{C4E4E3}
\definecolor{green1}{HTML}{A1D0C7}
\definecolor{green2}{HTML}{BFF6BA}
\definecolor{green3}{HTML}{028100}
\definecolor{teal}{HTML}{508AB2}
\definecolor{Gray}{gray}{0.94}
\definecolor{Gray2}{gray}{0.7}
\definecolor{orange3}{HTML}{c28c69}
\definecolor{blue3}{HTML}{3b75af}

\usepackage{pifont}

\usepackage{makecell}

\usepackage[listings]{tcolorbox}
\tcbuselibrary{skins, listings,theorems,breakable}
\newtcolorbox{mybox}{colback=white!5!white,colframe=black!75!black, left=.05in, right=.05in}

\newtcbtheorem{emptybox}{}%
{colback=green2!5,colframe=blue1,fonttitle=\bfseries,theorem style=plain, left=.0in, right=.0in,bottom=.02in, top=.02in,terminator sign none}{emptybox}
\newtcbtheorem[]{emptyboxx}{A Simple Template For Creating Backward Question}%
{colback=green2!5,colframe=blue1,fonttitle=\bfseries,theorem style=plain, left=.0in, right=.0in,bottom=.02in, top=.02in,terminator sign none}{emptyboxx}
\newtcbtheorem{bth}{Theoreme}
{colback=red!20,colframe=red,theorem style=plain,fonttitle=\bfseries,coltitle=black,terminator sign none}{th}

\definecolor{titlebg}{RGB}{0,0,86}        
\definecolor{framebg}{RGB}{0,0,86}        
\definecolor{bodybg}{RGB}{232,232,238}    

\newtcbtheorem[]{exmp}{Example}{
  colback=bodybg,
  colframe=framebg,
  coltitle=white,
  colbacktitle=titlebg,
  fonttitle=\bfseries,
  left=.02in,
  right=.02in,
  bottom=.02in,
  top=.02in,
  boxrule=1.2pt,
  arc=2mm,
  fontupper=\small\ttfamily,
  enhanced jigsaw, breakable
}{exmp}

\newtcbtheorem[]{template}{Template}%
{colback=green2!5,colframe=blue1,fonttitle=\bfseries, left=.02in, right=.02in,bottom=.02in, top=.02in}{template}
\newtcbtheorem[]{prompt}{Prompt}%
{colback=green2!5,colframe=blue1,fonttitle=\bfseries,
 left=.02in, right=.02in, bottom=.02in, top=.02in,
 fontupper=\small\ttfamily,
 enhanced jigsaw, breakable}{prompt}
\newtcbtheorem[number within=section]{thm}{Theorem}%
{colback=green!5,colframe=green!35!black,fonttitle=\bfseries}{th}
\newtcbtheorem[number within=section]{corr}{Corollary}%
{colback=green!5,colframe=green!35!black,fonttitle=\bfseries}{th}
\newtcbtheorem{method}{Method}%
{colback=green!5,colframe=green!35!black,fonttitle=\bfseries}{th}

\newtcbtheorem{ebox}{}
{colback=green2!5,colframe=blue1, left=.02in, right=.02in,bottom=.02in,title empty, top=.02in, theorem style=plain apart,frame empty}{ebox}
\theoremstyle{plain}

\theoremstyle{definition}

\theoremstyle{remark}

\newcommand{\appref}[1]{\hyperref[#1]{Appendix~\ref*{#1}}}

\begin{document}

\blfootnote{ \textsuperscript{\textdagger}Corresponding author}

\makeatletter
\renewcommand{\@notice}{}
\makeatother
\maketitle

\begin{abstract}
Inpatient medication recommendation requires clinicians to repeatedly select specific medications, doses, and routes as a patient's condition evolves. Existing benchmarks formulate this task as admission-level prediction over coarse drug codes with multi-hot diagnostic and procedure code inputs, failing to capture the per-timepoint, information-rich nature of real prescribing. 
We propose \methodName, a prescription-level benchmark that evaluates LLM prescribing capability by multiple-choice questions: each question presents a detailed patient profile and time-ordered clinical trajectory, requiring selection of specific medication-dose-route triples from real prescriptions and patient-specific distractors generated via reasoning-chain perturbation.
\methodName comprises 1{,}547 questions spanning 584 patients, 18 diagnostic categories, and 969 unique medications. Evaluation of 16 LLMs shows that \methodName is both challenging and discriminative: F1 ranges from 45.18 to 77.10 across models, and the best Exact Match is only 46.10\%. 
Error analysis reveals that even frontier models may overlook stated patient information and fail to derive clinical conclusions.
\end{abstract}


\section{Introduction}
\label{sec:introduction}

\footnote{*** revise the paper title. u should highlight some new features in the dataset // revised }
In inpatient care, clinicians repeatedly prescribe and adjust medications as a patient's condition evolves~\citep{chakraborty2014dynamic, viktil2012drug, rough2020predicting}. Each prescribing decision requires integrating laboratory results, vital-sign trends, treatment responses, and known contraindications to select the right medications, doses, and administration routes.
This task, known as medication recommendation~\citep{ali2023deep, wu2022conditional}, relies on deep specialist expertise.
However, the global shortage of trained clinicians leaves many hospitals under-resourced, particularly in critical-care and emergency settings, where patients may not receive appropriate medications~\citep{agyeman2023prioritising}.
Large language models (LLMs), which have demonstrated strong reasoning capabilities across medical domains~\citep{singhal2023large, nori2023capabilities}, offer a promising path toward automated medication recommendation, especially in expertise-constrained regions.
However, before any such system can be deployed, its realistic prescribing capability must be rigorously evaluated.

Many benchmarks (e.g., \citep{jin2021disease}) have been proposed to evaluate LLM medical capabilities, including licensing exams~\citep{pal2022medmcqa}, knowledge-based assessments~\citep{hendrycks2020measuring, gilson2023does}, and clinical tasks such as diagnosis~\citep{jin2019pubmedqa, arora2025healthbench}.
However, these benchmarks primarily evaluate medical knowledge rather than prescribing decisions, and frontier LLMs have begun to saturate them, exceeding 90\% accuracy on MedQA~\citep{jin2021disease}.

Recent works~\citep{shang2019gamenet, yang2021safedrug,  ma2024clibench, zhao2025finegrain, fan2025finegrained} have proposed an evaluation protocol for medication recommendation. 
They formulate the task as admission-level multi-label prediction, where each sample corresponds to an entire hospital admission, inputs are multi-hot ICD (International Classification of Diseases~\citep{who2021anatomical}) code vectors, \footnote{***what is ICD?// added} 
and outputs are ATC-3 (Anatomical Therapeutic Chemical~\citep{who2021anatomical}, level 3) therapeutic codes. \footnote{***what is ATC?//remove ATC.// added}
However, this evaluation protocol does not reflect real-world prescribing practice:
\begin{enumerate*}[(i), series = tobecont, itemjoin = \;]
\item it merges all prescriptions issued during an admission into a single label set, ignoring the fact that clinicians repeatedly adjust medications at different timepoints in response to evolving patient conditions.
\item multi-hot ICD input vectors discard the laboratory values, imaging findings, and clinical notes that clinicians rely on for individualized prescribing.
\item ATC-3 therapeutic codes collapse clinically distinct medications into a single label (e.g., heparin and enoxaparin share the same code despite different safety profiles) and discard dose and route, whereas real prescribing decisions require selecting specific medications with precise dosages and administration routes.
\end{enumerate*}

\begin{table}
  \vspace{-.1in}
  \centering
  \caption{Comparison with existing medication recommendation benchmarks.}
  \label{tab:benchmark_comparison}
  \footnotesize
  \setlength{\tabcolsep}{3pt}
  \begin{tabular}{l ccccc c}
    \toprule
    & \makecell{GAMENet \scriptsize\citep{shang2019gamenet}}
    & \makecell{SafeDrug \scriptsize\citep{yang2021safedrug}}
    & \makecell{CliBench \scriptsize\citep{ma2024clibench}}
    & \makecell{LAMO \scriptsize\citep{zhao2025finegrain}}
    & \makecell{FLAME \scriptsize\citep{fan2025finegrained}}
    & \methodName (ours) \\
    \midrule
    Prescription-level   & \ding{55} & \ding{55} & \ding{55} & \ding{55} & \ding{55} & \ding{51} \\
    Patient profile      & \ding{55} & \ding{55} & \ding{51} & \ding{51} & \ding{51} & \ding{51} \\
    Clinical trajectory  & \ding{55} & \ding{55} & \ding{55} & \ding{55} & \ding{55} & \ding{51} \\
    Specific medication        & \ding{55} & \ding{55} & \ding{55} & \ding{55} & \ding{55} & \ding{51} \\
    Dose \& route        & \ding{55} & \ding{55} & \ding{55} & \ding{55} & \ding{55} & \ding{51} \\
    LLM evaluation      & \ding{55} & \ding{55} & \ding{51} & \ding{51} & \ding{51} & \ding{51} \\
    \#medications & 145 & 131 & 182 & 151 & 151 & 969 \\
    \bottomrule
  \end{tabular}
  \vspace{-.2in}
\end{table}

To address these limitations, we propose \methodName, a benchmark for evaluating LLM prescribing capability in realistic inpatient scenarios.
\methodName is constructed from real electronic health records (EHR): each sample corresponds to a single prescribing timepoint rather than an entire admission, capturing the per-timepoint nature of real prescribing; 
its input includes a detailed patient profile and a time-ordered clinical trajectory, preserving the rich clinical information that clinicians rely on; 
and its output is a set of specific medication-dose-route triples rather than coarse therapeutic code (\autoref{tab:benchmark_comparison}).
Since evaluating open-ended generation is unreliable for prescribing decisions due to name variability and the existence of multiple clinically acceptable alternatives, we formulate each sample as a multiple-choice question (MCQ).
We propose a \emph{reasoning-chain perturbation} method to generate patient-specific distractors (incorrect options) by perturbing the clinical reasoning chain of correct medication, so that the question can only be answered through patient-specific clinical reasoning rather than general medical knowledge.

The resulting \methodName comprises 1{,}547 MCQs spanning 584 patients, 18 diagnostic categories, and 969 unique medications.
We evaluate 16 LLMs on \methodName and find that \methodName effectively discriminates LLMs of varying capabilities, with F1 score spanning 32 points (45.18 to 77.10).
Notably, GPT-4o scores only 58.90 F1 on \methodName despite achieving over 89\% on MedQA~\citep{jin2021disease}, and medical-domain fine-tuning yields negligible gains (at most $+2.20$ F1), confirming that \methodName evaluates capabilities beyond medical knowledge recall.
Even the strongest LLM (Gemini-3.1-Pro) achieves only 46.10\% Exact Match Accuracy, indicating that the task is far from solved.
Performance further degrades in late-phase admissions where longer clinical trajectories are integrated, suggesting that reasoning over extended temporal context remains a key challenge.
Case-study analysis reveals two common error patterns: LLMs either overlook patient information explicitly stated in the input (\emph{oversight errors}) or fail to derive clinical conclusions from the observations (\emph{reasoning errors}).

Our contributions are summarized as follows:
\begin{enumerate*}[(i), series = tobecont, itemjoin= \quad]
\item We formalize \emph{prescription-level} medication recommendation, shifting the task from admission-level prediction over coarse codes to per-timepoint decision making over detailed clinical trajectories and specific medication-dose-route triples.
\item We propose a \emph{reasoning-chain perturbation} method to generate patient-specific MCQ distractors, and construct \methodName, a benchmark of 1{,}547 questions spanning 584 patients, 18 diagnostic categories, and 969 unique medications.
\item 
We benchmark 16 LLMs and show that \methodName is highly discriminative and evaluates capabilities beyond medical knowledge recall. Even the strongest LLM achieves only 46\% Exact Match, and case-study analysis reveals systematic oversight and reasoning errors in frontier LLMs, highlighting significant room for improvement.\xfootnote{Due to page limit, a detailed discussion of related work is provided in~\appref{app:related_works}.}
\end{enumerate*}


\section{Prescription-level Medication Recommendation}
\label{sec:task_input}

\begin{figure}[!t]
    \centering
    \vskip -.1in
    \includegraphics[width=.95\linewidth]{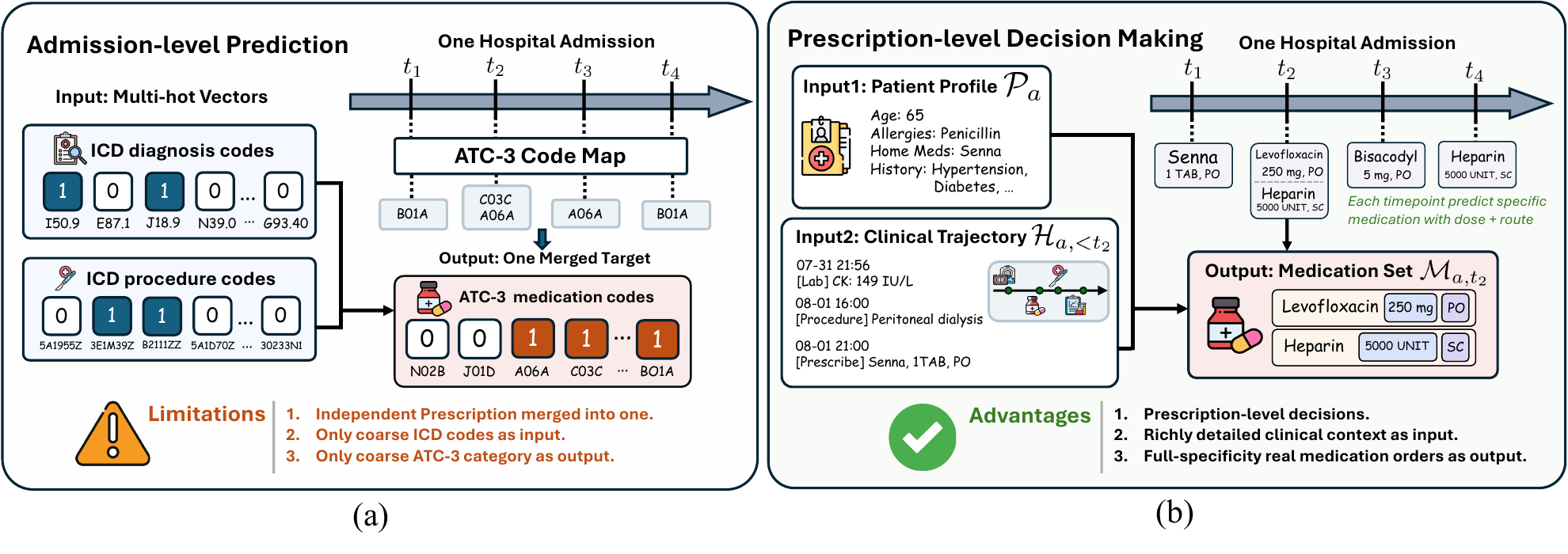}
    \vskip -.15in
    \caption{Comparison between admission-level and prescription-level medication recommendation. (a) Admission-level: all medications for an entire hospital admission are predicted at once, using only coarse diagnostic codes as input and coarse therapeutic-subclass codes as output. (b) Prescription-level: medications are predicted at each timepoint, based on the patient's detailed and evolving clinical state, with each output specifying the exact medication, dose, and route.}
    \label{fig:admission-vs-prescription}
    \vspace{-.15in}
\end{figure}

Existing benchmarks formulate medication recommendations as an \textit{admission-level} problem, predicting all medications corresponding to an entire hospital admission at once from only coarse diagnostic codes as input.
The formulation does not match how clinicians actually prescribe: decisions are made at specific timepoints, based on the patient's detailed and evolving clinical state.
In this section, 
we propose a \textit{prescription-level} medication recommendation formulation that reflects inpatient prescribing practice (\autoref{sec:formulation}) and describe how we construct the corresponding benchmark instances from electronic health records (EHR) data (\autoref{sec:data_sources}).
 
\subsection{From Admission-level Prediction to Prescription-level Decision Making}
\label{sec:formulation}

In inpatient care, medication prescribing is a dynamic and repeated process.
Over a \textit{single} hospital admission, clinicians issue prescriptions at multiple timepoints, each time reviewing the patient's detailed and changing clinical state, including symptom descriptions, laboratory results, vital-sign trends, imaging findings, and responses to prior treatments, to decide which medications to prescribe.
Because the patient's condition changes over time, prescriptions issued at different timepoints within the same admission often address very different clinical needs: in MIMIC-IV~\citep{johnson2023mimic}, a publicly available critical-care EHR database, the mean pairwise Jaccard similarity between medication sets prescribed at different timepoints within a single admission is only $0.054$.
Hence, a medication recommendation benchmark should capture the \emph{per-timepoint} nature of prescribing, the 
\emph{richly detailed} clinical information that informs 
each decision, and the \emph{full specificity} of real 
prescription orders.

However, existing benchmarks~\citep{shang2019gamenet, yang2021safedrug, li2024causalmed, liang2025cidgmed} meet none of these requirements.
These methods formulate the task as \emph{admission-level medication recommendation}, where each sample corresponds to an entire hospital admission.
The patient's clinical state is represented by multi-hot vectors over fixed vocabularies of ICD 
(International Classification of Diseases) 
diagnosis codes and ICD procedure codes~\citep{quan2005coding}
\xfootnote{ICD (International Classification of Diseases) is a standardized medical taxonomy that assigns a categorical label to each diagnosis (e.g., \texttt{I50.9} for ``heart failure, unspecified'') or procedure (e.g., \texttt{5A1955Z} for ``respiratory ventilation, greater than 96 hours'').}.
The prediction target is a multi-hot vector over a fixed 
vocabulary of ATC-3 
(Anatomical Therapeutic Chemical Classification, level 3) 
therapeutic codes~\citep{who2021anatomical},
\xfootnote{ ATC-3 (Anatomical Therapeutic Chemical Classification, level 3) is a pharmacological classification that groups medications by therapeutic subclass  (e.g., \texttt{C03C} ``high-ceiling diuretics'' covers furosemide, bumetanide, and torsemide alike).} 
and the label set is the union of all medications prescribed 
throughout the entire admission.
This formulation has four key problems:
\begin{enumerate*}[(i), series = tobecont, itemjoin = \;]
\item it aggregates all prescriptions issued during the admission into a single label set, merging the multiple, largely independent prescribing decisions described above into one prediction.
\item the inputs are limited to categorical ICD codes that carry no detailed clinical descriptions (such as the symptom narratives, lab values, imaging findings, and treatment responses) that clinicians rely on.
\item ATC-3 therapeutic codes operate at the subclass level, collapsing clinically distinct medications into a single label and obscuring safety-critical differences.\xfootnote{Heparin and enoxaparin share ATC-3 code B01A, yet only heparin is safe in severe renal impairment.} 
\item even when the correct medication is identified, ATC-3 codes discard the dose and administration route, both of which alter its clinical role.\xfootnote{Methotrexate at low oral doses treats rheumatoid arthritis, whereas at high intravenous doses it is chemotherapy.}
\end{enumerate*}

To address these limitations, 
we propose a \emph{prescription-level} task formulation: instead of one sample per admission, a single admission yields multiple samples, one per prescribing timepoint.
Let $a$ index an admission and $t$ a prescribing timepoint within $a$.
For each $(a, t)$, the task is to predict the medication orders $\mathcal{M}_{a,t}$ issued at $t$, given a patient profile $\mathcal{P}_{a}$ and clinical trajectory $\mathcal{H}_{a,<t}$ before $t$:
\begin{equation}
\mathcal{F}: (\mathcal{P}_{a},\, \mathcal{H}_{a,<t}) \;\mapsto\;
\mathcal{M}_{a,t}.
\label{eq:med_prompt_goal}
\end{equation}
The patient profile 
$\mathcal{P}_{a}$ is a \emph{detailed textual description} of the patient's pre-admission state, including chronic conditions, drug allergies, and home medications, providing the baseline context that applies throughout the admission.
The clinical trajectory $\mathcal{H}_{a,<t}$ is a time-ordered sequence of clinical events occurring before $t$ within the current admission, including symptom notes, laboratory results, vital-sign measurements, and prior medication orders, to describe how the patient's condition has evolved up to the prescribing moment.
Together, we replace the coarse categorical ICD codes used in existing benchmarks with the richly detailed clinical information that clinicians actually rely on.

The output $\mathcal{M}_{a,t}$ is a \emph{set} of medication orders issued at timepoint $t$, where each element is a $(\textit{medication},\, \textit{dose},\, \textit{route})$ triple specifying the medication, its dosage, and its administration route (e.g., oral, intravenous).
Because a clinician may issue multiple medication orders
at a single prescribing moment, the model must both identify the correct medications and determine how many to prescribe.
Next, we describe how we instantiate $\mathcal{P}_{a}$, $\mathcal{H}_{a,<t}$, and $\mathcal{M}_{a,t}$ from the EHR data.

\subsection{From EHR Data to Benchmark Instances}
\label{sec:data_sources}

We now describe how we instantiate $\mathcal{P}_{a}$, $\mathcal{H}_{a,<t}$, and $\mathcal{M}_{a,t}$ from electronic health records (EHR).
All data used below, including demographics, clinical notes, lab results, radiology reports, and medication orders, are routinely recorded during hospital admissions and are readily available from MIMIC-IV~\citep{johnson2023mimic}, a publicly available critical-care EHR database. 

\textbf{Patient profile $\mathcal{P}_{a}$.}
Different medications are appropriate for different patient backgrounds. A medication that is safe for one patient may be harmful for another due to a chronic condition, a known allergy, or an interaction with a home medication.
To capture these constraints, $\mathcal{P}_{a}$ combines structured demographics (age, sex, admission type) with textual descriptions of past medical history, allergies, and home medications extracted from clinical notes.
Because these prescribing-relevant backgrounds are typically documented only in free-text notes, we use GPT-5-Chat~\citep{singh2025openai} to extract them, instructing it to retrieve only content documented as true \emph{prior to admission} and to output a fixed set of section headings for consistency across patients. The extraction prompt is in ~\appref{app:prompts}.

\textbf{Clinical trajectory $\mathcal{H}_{a,<t}$.}
Prescribing decisions depend not only on the patient's profile but also on how their condition has evolved since admission: a rising lab value, a new imaging finding, or a lack of response to a prior medication can each change what should be prescribed next.
To provide this evolving context, $\mathcal{H}_{a,<t}$ combines four types of time-stamped intra-admission events before $t$:
\begin{itemize}[leftmargin=1.2em, itemsep=0pt, topsep=2pt]
\item \textbf{Prescriptions}: 
medication name, dose, and administration route, recording prior orders.
\item \textbf{Lab results}: test name, value, and unit, reflecting the patient's changing physiology.
\item \textbf{Procedures}: procedure descriptions, indicating interventions already performed.
\item \textbf{Radiology reports}: the \textsc{Findings} and \textsc{Impression} sections of MIMIC-IV radiology notes,
providing imaging-based assessments.
\end{itemize}
Each event corresponds to a single recorded entry (a lab test, a prescription order, a procedure, or a radiology report). Events sharing the same timestamp are merged into a single block, and blocks are concatenated in chronological order to form a free-text timeline that preserves the order in which information became available to the clinician (\hyperref[fig:admission-vs-prescription]{Figure~\ref*{fig:admission-vs-prescription}(b)}). 
An example is shown in \appref{app:examples}.

\textbf{Prescription set $\mathcal{M}_{a,t}$}
is the set of medications prescribed at timepoint $t$ in admission $a$, where each element is a $(\textit{medication},\, \textit{dose},\, \textit{route})$ triple that specifies what medication to give, how much, and how to administer it.
We keep the original medication names as documented in the EHR, 
preserving the full specificity of real inpatient prescribing 
rather than mapping to coarse therapeutic subclasses as in prior 
benchmarks~\citep{yang2021safedrug, zhao2025finegrain, 
fan2025finegrained}.

\section{Constructing Multiple-Choice Questions via Reasoning-Chain Perturbation}
\label{sec:eval_design}
 
Given the prescription-level task defined in \autoref{sec:task_input}, we now describe how we construct the evaluation.
We first explain why we use MCQs over open-ended generation (\autoref{sec:eval_setup}), then describe how we construct patient-specific distractors (incorrect options) by perturbing clinical reasoning chains (\autoref{sec:distractor_generation}), and finally assemble the complete MCQs (\autoref{sec:assembly}).
For each admission, we select timepoints that represent active prescribing decisions (excluding routine and continued orders) and balance across early, middle, and late admission phases (details in ~\appref{app:timepoint_selection}).
 
\subsection{Evaluation via Multiple-Choice Questions}
\label{sec:eval_setup}

A direct way to evaluate medication recommendation is to have LLMs freely generate medications given $(\mathcal{P}_a, \mathcal{H}_{a,<t})$ and compare the output against $\mathcal{M}_{a,t}$.
However, this open-ended setup is difficult to evaluate reliably.
Many prescribing decisions admit multiple clinically acceptable answers (e.g., metoprolol versus carvedilol for heart failure), so an LLM that outputs a reasonable but non-matching medication would be incorrectly penalized.
Additionally, medication names vary widely across abbreviations and formulations, making evaluation unreliable.

We therefore cast each prediction as a \textbf{multiple-choice question (MCQ)}:
the LLM is presented with the patient context $(\mathcal{P}_a, \mathcal{H}_{a,<t})$ together with a set of candidate medications as options, and is asked to prescribe medications from these options.
This avoids both problems above: the correct medications are explicitly listed among the options, so scoring is unambiguous; and the LLM selects from a fixed set rather than generating free-form text, eliminating name variability.

The key challenge in constructing MCQs is generating the incorrect options (distractors).
The correct options are simply the medications in $\mathcal{M}_{a,t}$, but the distractors must be carefully designed: they should be \emph{wrong} for this patient at this timepoint, yet \emph{clinically related} to the patient's condition, so that distinguishing them from the correct answers requires clinical reasoning.
In the next section, we describe how we generate such patient-specific distractors.

\subsection{Distractor Construction via Reasoning-Chain Perturbation}
\label{sec:distractor_generation}

\begin{figure}[t]
    \centering
    \vskip -.1in
    \includegraphics[width=.95\linewidth]{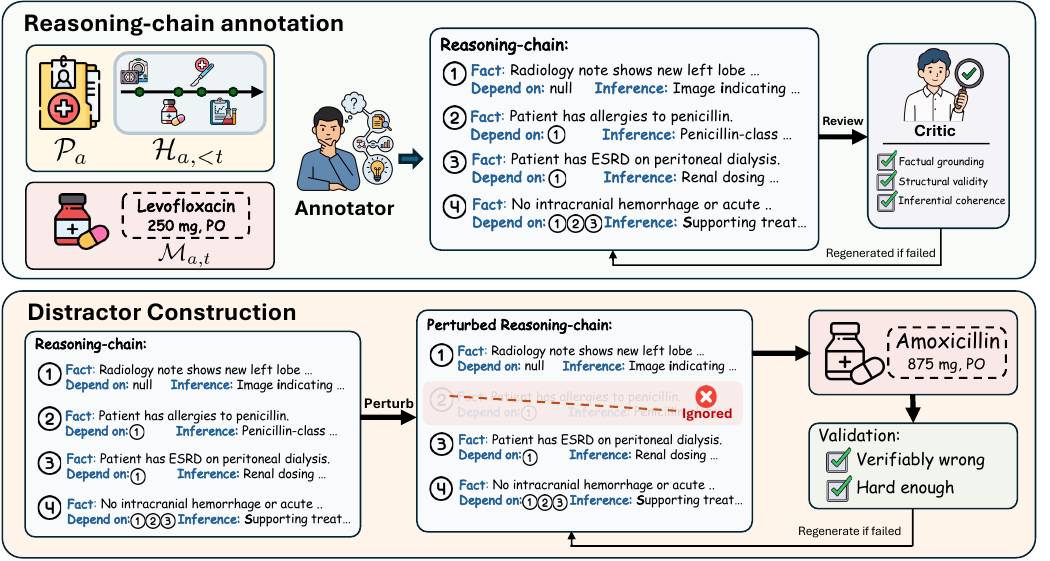}
    \caption{Overview of \methodName MCQ construction pipeline. Reasoning-chain annotation (top) extracts stepwise clinical reasoning for each correct medication and validates it through a critic that checks factual grounding and logical coherence. Distractor generation and validation (bottom) produces patient-specific distractors by perturbing the reasoning chain and verifies that each distractor is both verifiably wrong and sufficiently difficult. Failed chains and distractors trigger iterative revision.}
    \label{fig:pipeline}
    \vspace{-.2in}
\end{figure}

As discussed above, distractors must be wrong for this patient at this timepoint, yet clinically related to the patient's condition, so that ruling them out requires reasoning over the patient's actual clinical context.
To achieve this, we first use an LLM to annotate a reasoning chain that explains why a correct medication is prescribed for this patient, then perturb steps in that chain to produce distractors that would be appropriate under slightly different conditions but are inappropriate for this patient.
As illustrated in \autoref{fig:pipeline}, this is done through three steps.

\textbf{Step 1: Reasoning-chain annotation.}
For each correct medication $m \in \mathcal{M}_{a,t}$, we use an annotator LLM to annotate a \emph{reasoning chain} that explains why $m$ is prescribed for this patient at this timepoint.
The chain breaks down the prescribing rationale into a sequence of steps, each linking a clinical observation from the patient's record to a clinical inference.
\xfootnote{For example, in \autoref{fig:pipeline}, the chain for Levofloxacin starts with imaging showing left lower lobe consolidation, which suggests pneumonia and establishes the need for an antibiotic; subsequent steps then narrow the medication choice by noting the patient's penicillin allergy (ruling out $\beta$-lactams) and end-stage renal disease (requiring dose adjustment).}
To ensure that the chain is faithful to the patient's record, a critic LLM verifies that every cited observation actually appears in $\mathcal{H}_{a,<t}$ and that the reasoning is logically coherent.
Chains that fail this check are regenerated up to three times before the timepoint is discarded.

\textbf{Step 2: Distractor generation via perturbation.}
Given a reasoning chain, we use an LLM first to perturb reasoning steps, either ignoring a clinical observation or changing the condition it establishes, and then to prescribe a medication based on the perturbed chain.
The resulting distractor is a medication that would be appropriate under the perturbed reasoning chain but is inappropriate given the patient's original clinical context.
\xfootnote{For example, if we ignore the step related to penicillin allergy in the chain above, amoxicillin-clavulanate becomes a natural choice: it is a standard first-line antibiotic for aspiration pneumonia, but it is wrong for this patient due to theallergy.}
Because the perturbed chain is still derived from this patient's clinical record, the resulting distractor remains related to the patient's condition, making it difficult to rule out without carefully examining the patient's record.

\textbf{Step 3: Distractor validation.}
Each candidate distractor is checked along two dimensions:
\begin{enumerate*}[(i), series = tobecont, itemjoin = \;]
\item  \emph{is it verifiably wrong?}
There must be specific evidence in the patient's record that proves the distractor is inappropriate.
For example, if the rationale claims ``the distractor medication is wrong because the patient has a penicillin allergy,'' the allergy must be documented in the record.
\item \emph{is it hard enough?}
Ruling out the distractor requires the patient's specific clinical information.
For example, amoxicillin-clavulanate passes this check because it can only be ruled out by knowing this patient's penicillin allergy; nitrofurantoin fails because it does not treat pneumonia regardless of the patient.
Distractors that fail either check are discarded and regenerated up to three times.
\end{enumerate*}

\textbf{Physician validation.}
To verify that the generated distractors are indeed inappropriate for the target patient, a licensed physician reviewed 95 distractors from 19 prescription timepoints, judging whether each medication should be prescribed for the given patient.
96.8\% were judged as clearly inappropriate, 3.2\% as clinically debatable, and none as appropriate, confirming that the pipeline reliably produces verifiably wrong distractors.

\subsection{Assembling MCQs}
\label{sec:assembly}

For each timepoint that passes all validation steps, we combine the correct medications in $\mathcal{M}_{a,t}$ with five distractors generated in \autoref{sec:distractor_generation}, shuffle the combined set, and label the options with alphabetic letters.
The question presents the patient profile $\mathcal{P}_a$ followed by the clinical trajectory $\mathcal{H}_{a,<t}$, and asks the LLM to prescribe medications from the listed options.
An example is shown in 
\appref{app:examples}.
We use the GPT-5 series~\citep{singh2025openai} for all LLM-assisted steps in the pipeline.
All prompts used in the pipeline are provided in~\appref{app:construct_mcq}.

\section{Benchmark Dataset RxEval}
\label{sec:dataset_analysis}

\begin{figure}[t]
\centering
\begin{subfigure}[t]{0.26\textwidth}
    \vspace{0pt}
    \centering
    \scriptsize
    \renewcommand{\arraystretch}{0.9}
    \setlength{\tabcolsep}{2pt}
    \resizebox{\textwidth}{!}{
    \begin{tabular}[t]{@{}lr@{}}
      \toprule
      \# Questions & 1{,}547 \\
      \# Admissions & 587 \\
      \# Unique patients & 584 \\
      \# ICD chapters & 18 \\
      Avg Q.\ / admission & 2.64 \\
      \midrule
      \# Unique medications & 969 \\
      \quad Correct & 436 \\
      \quad Distractor & 740 \\
      \midrule
      Correct opt.\ (avg/max) & 2.03 / 18 \\
      Total opt.\ (avg/max) & 7.03 / 23 \\
      Tokens (med./max) & 3{,}111 / 118{,}805 \\
      \midrule
      Phase E/M/L (\%) & 31.4 / 32.9 / 35.7 \\
      \bottomrule
    \end{tabular}
    }
    \caption{Summary statistics}
    \label{tab:basic_stats}
\end{subfigure}\hfill
\begin{subfigure}[t]{0.40\textwidth}
    \vspace{0pt}
    \centering
    \includegraphics[width=\textwidth]{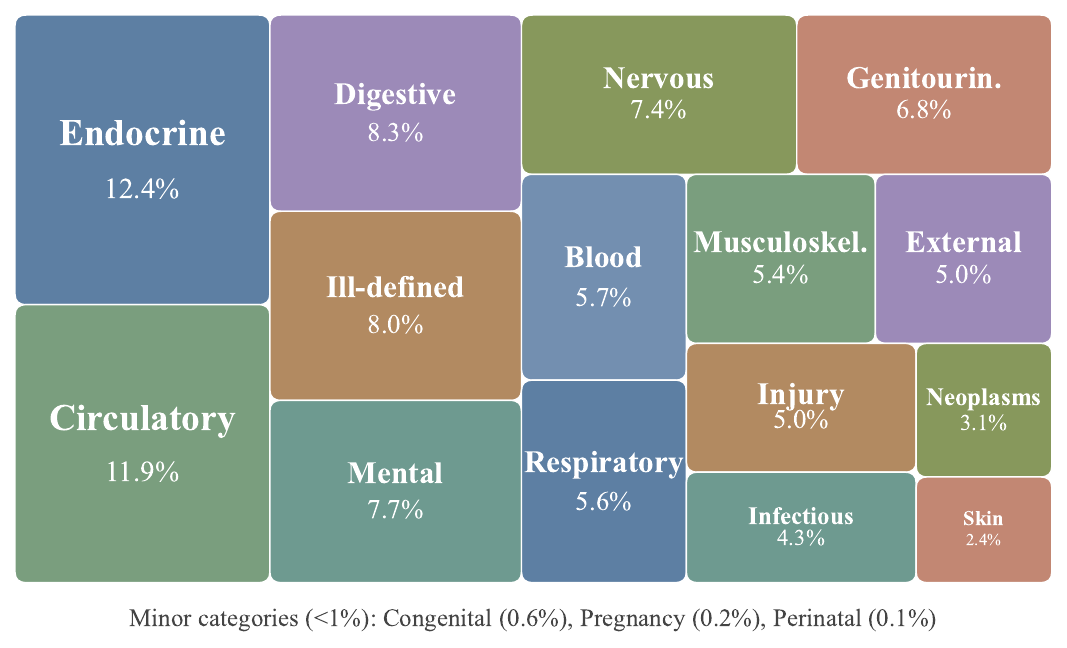}
    \caption{ICD chapter distribution}
    \label{fig:icd_chapter}
\end{subfigure}\hfill
\begin{subfigure}[t]{0.30\textwidth}
    \vspace{0pt}
    \centering
    \includegraphics[width=\textwidth]{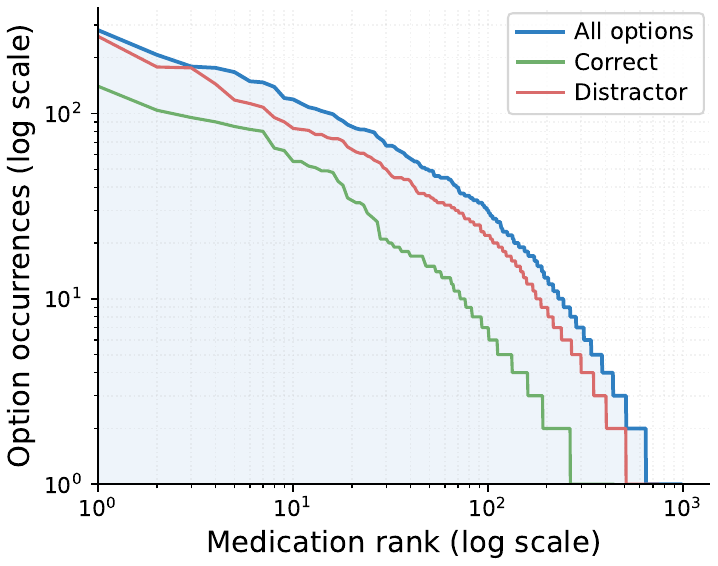}
    \caption{Medication frequency}
    \label{fig:med_freq}
\end{subfigure}

\vspace{0.5em}

\begin{subfigure}[t]{0.31\textwidth}
    \centering
    \includegraphics[width=\textwidth]{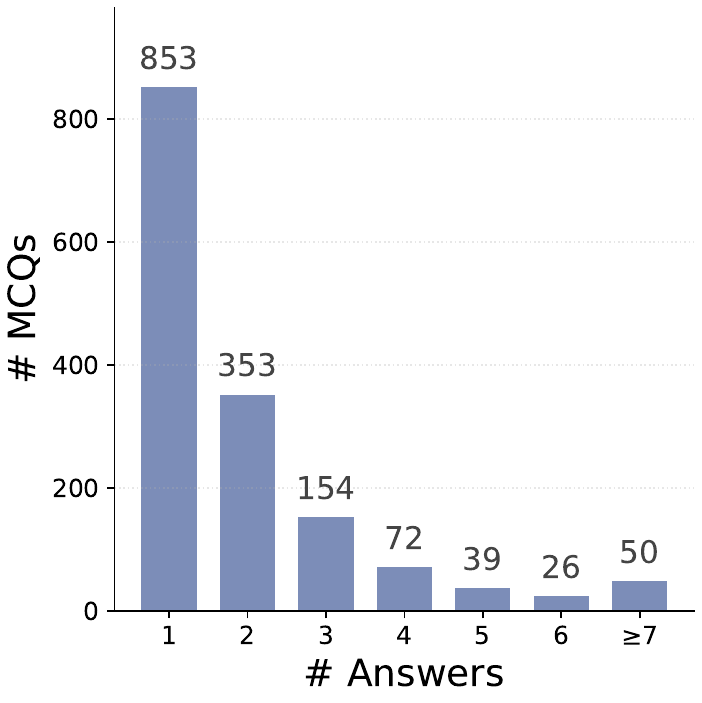}
    \caption{Answers per MCQ}
    \label{fig:answers_per_mcq}
\end{subfigure}\hfill
\begin{subfigure}[t]{0.31\textwidth}
    \centering
    \includegraphics[width=\textwidth]{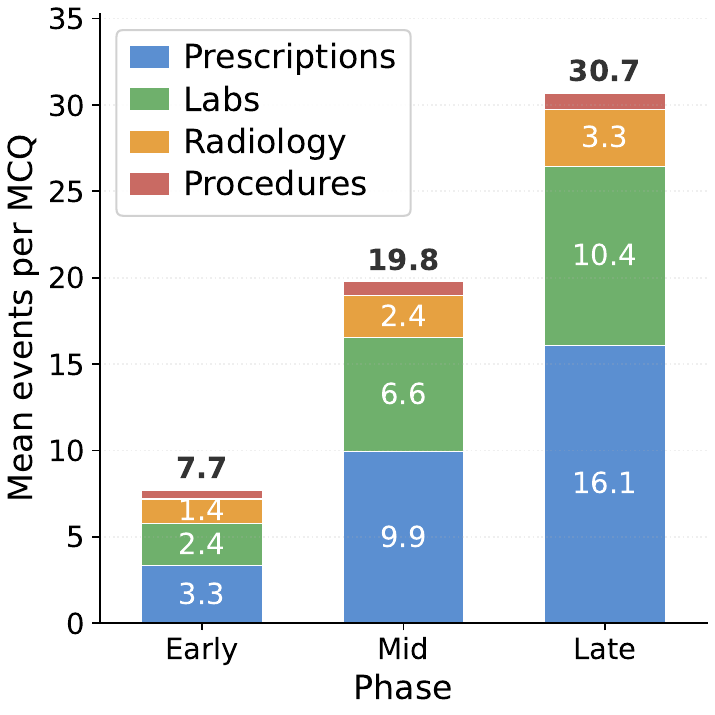}
    \caption{Events by phase}
    \label{fig:traj_events}
\end{subfigure}\hfill
\begin{subfigure}[t]{0.36\textwidth}
    \centering
    \includegraphics[width=\textwidth]{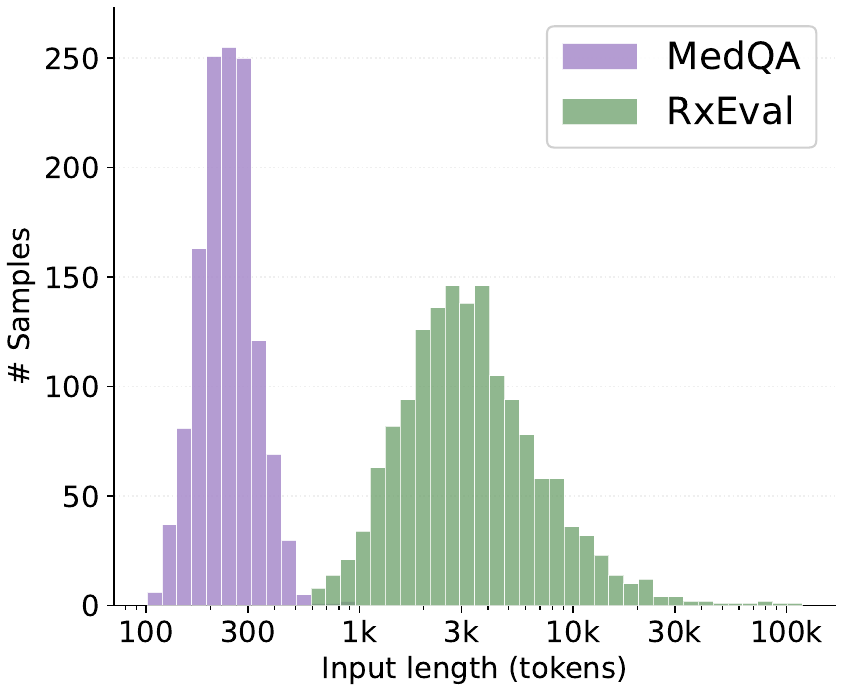}
    \caption{Input token length}
    \label{fig:token_length}
\end{subfigure}

\caption{Dataset composition of \methodName. (a)~Summary statistics. (b)~Diagnostic coverage across 18 ICD chapters. (c)~Long-tail medication frequency for correct and distractor medications. (d)~Distribution of correct options per MCQ: 44.9\% are multi-answer. (e)~Mean clinical events per MCQ by temporal phase, showing richer context in later stages. (f)~Prompt token length distribution.}
\label{fig:dataset_stats}
\vspace{-.2in}
\end{figure}

Our constructed \methodName comprises 1{,}547 MCQs drawn from 587 admissions
of 584 unique patients in MIMIC-IV, curated from an initial pool of 11{,}257 candidate prescription timepoints through
successive stages of sampling, timepoint selection, reasoning-chain annotation, and distractor validation.
\autoref{tab:basic_stats} summarizes the key statistics.
Below, we describe two strengths of the benchmark: broad clinical
coverage and challenging task design.

\textbf{Broad clinical coverage.}
A medication benchmark should reflect the diversity of real inpatient care. The inpatient cohort has a mean age of 59.5 years (range 18--91, 53.7\% female) and covers emergency (48\%), observation (27\%), surgical same-day (11\%), urgent (10\%), and elective (4\%) admissions.
Diagnoses span all 18 ICD-9 chapter-level categories (\autoref{fig:icd_chapter}), led by Circulatory (21.1\%),
Injury (12.1\%), and Digestive (11.3\%), matching the case mix of the MIMIC-IV population. The medication space contains 969 unique medications with a long-tail distribution
(\autoref{fig:med_freq}): most medications appear in fewer than five MCQs. 
This means LLMs cannot rely on memorizing common prescriptions and must generalize to infrequent medications.

\textbf{Challenging task design.}
\methodName differs from standard four-option single-answer medical MCQs \citep{jin2021disease,pal2022medmcqa}
in three key dimensions: 
(i) each question has 7.03 options on
average (median 6, range 5--23), and the correct set ranges from 1 to 18 (mean 2.03), with 44.9\% of questions being multiple-answer (\autoref{fig:answers_per_mcq}). Models must therefore select the correct medication subset without a fixed target size. 
(ii) the clinical context grows as admissions progress, from a mean of 7.7 events per question in the early phase to 30.7 in the late phase (\autoref{fig:traj_events}). Late-phase questions require models to reason over longer and more complex histories. Questions are balanced across the early (31.4\%), middle (32.9\%), and late (35.7\%) phases to ensure evaluation covers all difficulty levels.
(iii) the resulting prompt length has a median of 3{,}111 tokens (P95 = 12{,}504, \autoref{fig:token_length}), with complex admissions exceeding 100K tokens. This is roughly $13\times$ longer than MedQA (median 239 tokens).

\section{Experiments}
\label{sec:experiments}

\subsection{Experimental Setup}
\label{sec:setup}
 
\textbf{LLMs.}
We evaluate 16 LLMs spanning eight proprietary and eight open-source models.
The proprietary set includes models from the GPT and Gemini families; the open-source set includes general-purpose models of varying scales (8B--72B) and three medical-domain models fine-tuned from the corresponding base checkpoints, enabling controlled comparison of medical fine-tuning versus general instruction tuning.
Full model details are provided in~\appref{app:models}. 
We additionally include a Random baseline
that independently selects each option with equal probability ($p{=}0.5$), establishing chance-level performance without any dataset-specific knowledge.

\textbf{Evaluation metrics.}
Let $\hat{Y}$ and $Y$ denote the predicted and correct option sets for a question.
Following~\citep{fan2025finegrained},
we report five metrics averaged over all questions:
\begin{enumerate*}[(i), series = tobecont, itemjoin = \quad]
\item \textbf{Exact Match score}
is 1 if $\hat{Y}=Y$ and 0 otherwise;
    \item \textbf{Jaccard similarity}${} = |\hat{Y}\cap Y|/|\hat{Y}\cup Y|$;
  \item   
  \textbf{Precision}${} = |\hat{Y}\cap Y|/|\hat{Y}|$;
  \item \textbf{Recall}${} = |\hat{Y}\cap Y|/|Y|$;
  \item \textbf{F1 score}${} = 2\cdot\text{Precision}\cdot\text{Recall}/(\text{Precision}+\text{Recall})$.
\end{enumerate*}

\textbf{Implementation details.}
All 1{,}547 MCQs in \methodName serve as the test set.
All LLMs are evaluated in a zero-shot setting with temperature 0. For proprietary LLMs, we use the official APIs with the model versions. For open-source LLMs, we deploy them using vLLM on NVIDIA A100 80GB GPUs with bf16 precision.
Prompts exceeding a model's context window are truncated by removing the earliest trajectory events while preserving the patient profile and most recent clinical context.

 \subsection{Main Results}
\label{sec:main_results}

\autoref{tab:mcq_results}
reports the main results across 16 LLMs and we highlight 
the following
four findings.

\textbf{\methodName provides strong discriminative power.}
F1 spans nearly 32 points across all 16 LLMs (77.10 for Gemini-3.1-Pro to 45.18 for Llama-3.1-8B-Instruct), confirming that \methodName effectively separates LLMs of different capabilities.
However, Exact Match is far more demanding: Gemini-3.1-Pro achieves only 46.10\% Exact Match despite 77.10 F1, while Llama-3.1-8B-Instruct attains 45.18 F1 yet its Exact Match of 1.36\% is barely above the Random baseline (1.08\%).
This Exact Match--F1 gap reveals that weaker models can identify individually plausible medications but seldom assemble a fully correct prescription set.
The difficulty stems from three design choices: the input is a lengthy, richly detailed clinical trajectory, the multi-answer format requires identifying a correct medication subset whose size varies across questions, and the reasoning-chain perturbation (\autoref{sec:distractor_generation}) produces patient-specific distractors that cannot be ruled out by general medical knowledge alone.

\begin{figure}[!b]
\centering
\vskip -.15in
\begin{minipage}[t]{0.33\linewidth}
    \centering
    \includegraphics[width=\linewidth]{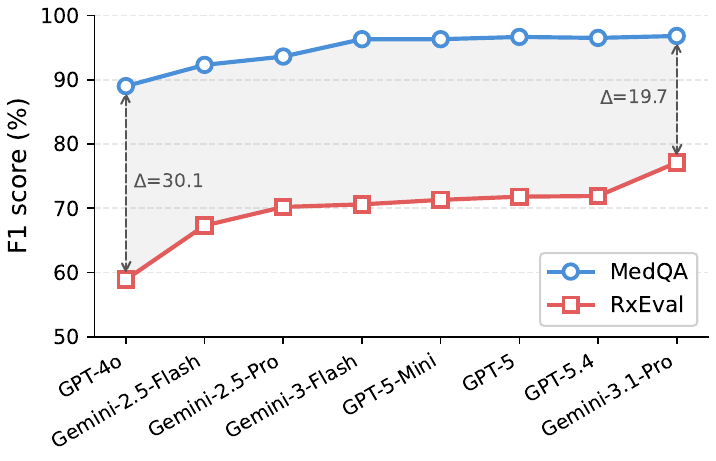}
    \vskip -.1in
    \caption{\methodName vs MedQA across proprietary models.}
    \label{fig:medqa_vs_temrx}
\end{minipage}
\hfill
\begin{minipage}[t]{0.28\linewidth}
    \centering
    \includegraphics[width=\linewidth]{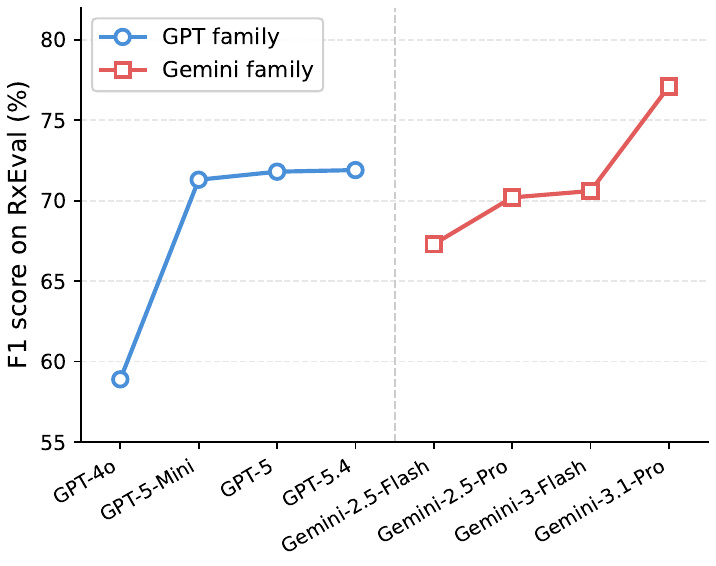}
    \vskip -.1in
    \caption{F1 of GPT and Gemini families on \methodName.}
    \label{fig:family_timeline}
\end{minipage}
\hfill
\begin{minipage}[t]{0.32\linewidth}
    \centering
    \includegraphics[width=\linewidth]{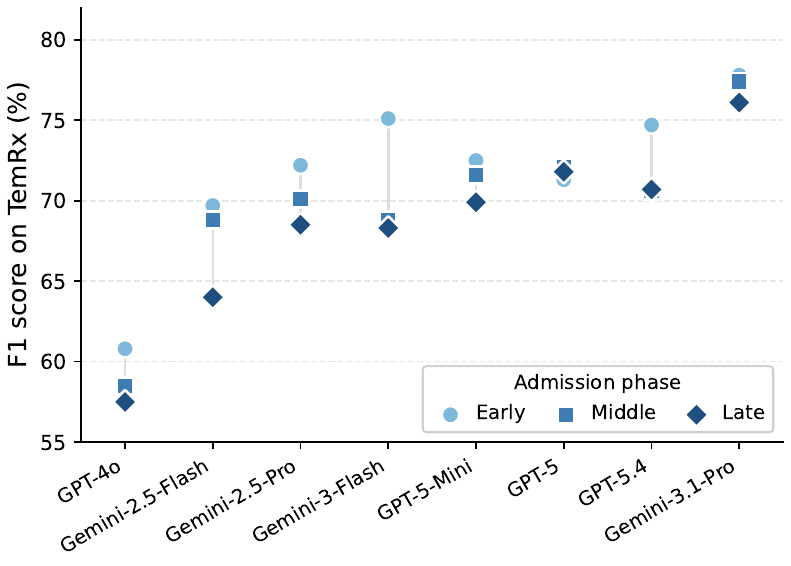}
    \vskip -.1in
    \caption{F1 by admission phase across models.}
    \label{fig:temporal_phase}
\end{minipage}
\end{figure}

\textbf{\methodName evaluates temporal clinical reasoning, not simple medical knowledge recall.}
\autoref{fig:medqa_vs_temrx} compares the closed-source models' F1 scores 
on \methodName against their MedQA~\citep{jin2021disease} performance as reported on the VALS
leaderboard~\cite{vals2025medqa}.
All eight models score above 89 on MedQA, indicating that MedQA
is approaching saturation for frontier LLMs.
On \methodName, the same models exhibit a much wider performance
spread (58.9--77.1 F1), suggesting that \methodName probes
capabilities beyond medical knowledge recall:
while MedQA tests factual knowledge via short single-answer questions, \methodName requires integrating a
patient's evolving clinical trajectory to assemble a multi-medication
prescription (see \autoref{fig:case_study} in \autoref{app:case_study} for examples).

\begin{table*}[t]
\centering
\small
\caption{MCQ evaluation results on \methodName, ranked by F1. Best overall is \textbf{bold}; best among open-source models is \underline{underlined}. 
The medical-domain models are marked with
$^\dagger$.}
\vskip -.1in
\label{tab:mcq_results}
\begin{NiceTabular}{cc|ccccc}
\toprule
 & Model & Exact Match & Jaccard & Precision & Recall & F1 \\
\midrule
\Block{8-1}{\rotate \textit{Proprietary LLM}}
 & Gemini-3.1-Pro       & 46.10          & \textbf{69.90} & 76.50          & \textbf{84.60} & \textbf{77.10} \\
 & GPT-5.4              & 47.90          & 66.40          & 79.20          & 70.30          & 71.90 \\
 & GPT-5                & \textbf{48.70} & 66.30          & \textbf{80.40} & 69.00          & 71.80 \\
 & GPT-5-Mini           & 42.70          & 64.60          & 76.50          & 72.30          & 71.30 \\
 & Gemini-3-Flash       & 36.10          & 62.20          & 67.10          & 85.80          & 70.60 \\
 & Gemini-2.5-Pro       & 37.60          & 62.40          & 71.10          & 76.50          & 70.20 \\
 & Gemini-2.5-Flash     & 38.00          & 59.90          & 70.40          & 75.90          & 67.30 \\
 & GPT-4o               & 21.70          & 49.50          & 57.90          & 68.30          & 58.90 \\
\midrule
\Block{8-1}{\rotate \textit{Open LLM}}
 & Qwen3.6-27B          & \underline{40.00} & \underline{64.20} & \underline{74.20} & 75.30 & \underline{71.70} \\
 & HuatuoGPT-o1-72B$^\dagger$     & 10.30 & 44.90 & 50.70 & \underline{78.50} & 57.30 \\
 & MedGemma-27B$^\dagger$         & 16.70 & 46.70 & 55.30 & 68.50 & 57.20 \\
 & Qwen2.5-72B-Instruct &  9.90 & 44.00 & 49.80 & 75.20 & 56.10 \\
 & Gemma-3-27B-IT       &  9.80 & 43.20 & 50.70 & 68.90 & 55.00 \\
 & Qwen2.5-7B-Instruct  & 10.10 & 42.40 & 49.80 & 65.70 & 53.00 \\
 & HuatuoGPT-o1-8B$^\dagger$      &  2.71 & 33.62 & 39.02 & 71.70 & 46.51 \\
 & Llama-3.1-8B-Instruct &  1.36 & 32.20 & 36.72 & 71.48 & 45.18 \\
\midrule
 & Random           &  1.08 & 21.57 & 26.66 & 50.00 & 31.26 \\
\bottomrule
\end{NiceTabular}
\vspace{-.15in}
\end{table*}

\textbf{Medical-domain fine-tuning does not transfer to \methodName.}
The three medically fine-tuned
models in our evaluation are trained predominantly on medical QA and clinical knowledge probing tasks~\citep{chen2024huatuogpt,sellergren2025medgemma}, yet they offer negligible gains on \methodName: 
HuatuoGPT-o1-72B exceeds its general instruction-tuning counterpart Qwen2.5-72B-Instruct
by only 1.20 F1 (57.30 vs.\ 56.10), MedGemma-27B leads Gemma-3-27B-IT by 2.20 (57.20 vs.\ 55.00), and HuatuoGPT-o1-8B exceeds Llama-3.1-8B-Instruct by 1.33 (46.51 vs.\ 45.18).
This minor transfer confirms that the temporal clinical reasoning required by \methodName is not covered by existing medical-domain fine-tuning paradigms.

\textbf{\methodName tracks frontier progress.}
\autoref{fig:family_timeline}
traces the F1 progression of the GPT and Gemini families on \methodName.
Within each family, generational upgrades yield substantial gains: GPT-4o\,$\to$\,GPT-5 brings $+12.90$ F1, and Gemini-2.5-Pro\,$\to$\,Gemini-3.1-Pro brings $+6.90$, confirming that
\methodName is sensitive to the capability gains introduced
by successive frontier releases.
 
\subsection{Error Analysis}
\label{sec:failure_analysis}
 
We further analyze model behavior along three dimensions: (i) difficulty across admission phases, (ii) the effect of answer-set size, and (iii) recurring error types in the strongest models.


\textbf{Late-phase prescriptions are harder.}
We split the RxEval questions
into early, middle, and late phases of each admission and report the F1 scores by phase in \autoref{fig:temporal_phase}.
Six of the eight models
reach their lowest F1's in the late phase, where the accumulated trajectory is longest and contains the most clinical events (\autoref{fig:traj_events}). 
The drop is largest for mid-tier models such as Gemini-2.5-Flash ($-5.70$ F1) and Gemini-3-Flash ($-6.80$), while frontier models drop less (Gemini-3.1-Pro $-1.70$; GPT-5 essentially flat),confirming that integrating longer temporal context is more difficult in \methodName.

\begin{wrapfigure}{r}{0.35\linewidth}
\centering
\vspace{-0.15in}
\includegraphics[width=\linewidth]{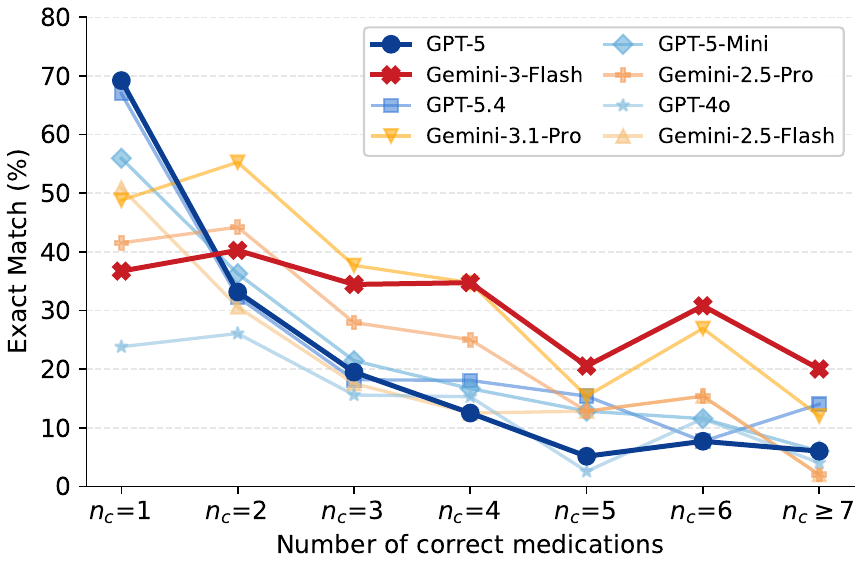}
\vskip -.1in
\caption{Exact-match score by number of ground-truth medications ($n_c$).}
\label{fig:em_by_nc}
\vspace{-.15in}
\end{wrapfigure}

\textbf{Larger answer sets are harder.}
\autoref{fig:em_by_nc} shows Exact Match w.r.t. the number of ground-truth medications $n_c$.
All models score substantially lower on large-$n_c$
questions than on single-answer ones.
For example, GPT-5 drops from 69.2\% Exact Match at $n_c{=}1$
to 6.0\% at $n_c{\geq}7$.
Gemini-3-Flash exhibits a gentler decline (36.7\% to 20.0\%),
yet still struggles as $n_c$ grows.
No model exceeds 20\% Exact Match when $n_c{\geq}7$,
indicating that predicting the correct multi-medication subset
becomes more difficult as the label size grows.

\textbf{Case study: Oversight errors and reasoning errors.}
The preceding analyses identify \emph{when} models struggle; we now examine \emph{why} by analyzing representative failures from GPT-5.4 and Gemini-3.1-Pro (\autoref{fig:case_study} in \appref{app:case_study}). 

\emph{Oversight errors} occur when the model ignores evidence plainly stated in the input.
In \hyperref[fig:case_study]{Figure~\ref*{fig:case_study}(a)}, GPT-5.4 prescribes Enoxaparin despite a documented heparin allergy; in \hyperref[fig:case_study]{Figure~\ref*{fig:case_study}(b)}, Gemini-3.1-Pro prescribes Dextrose 50\% for a patient whose glucose is already elevated at 151 mg/dL.
These errors require no reasoning to avoid.
Indeed, when the allergy list is repeated at the end of the prompt, 85.4\% of the allergy-violation cases can be corrected (Details in \appref{app:allergy_repeat}).

\emph{Reasoning errors} occur when the model fails to derive a clinical conclusion from observations in the input.
In \hyperref[fig:case_study]{Figure~\ref*{fig:case_study}(c)}, GPT-5.4 prescribes renally cleared IV Magnesium Sulfate despite elevated creatinine (2.3 mg/dL), failing to infer impaired renal function; in \hyperref[fig:case_study]{Figure~\ref*{fig:case_study}(d)}, Gemini-3.1-Pro prescribes IV Labetalol for elevated BP that is caused by alcohol withdrawal rather than primary hypertension, failing to attribute the observation to its underlying cause.

These two failure modes point to complementary gaps: models do not reliably use patient-specific information present in the input (oversight), and do not reliably combine clinical observations with medical knowledge (reasoning).

\section{Conclusion}
We introduced \methodName, a benchmark for evaluating LLM
prescribing capability in realistic inpatient scenarios. By shifting
from admission-level prediction over coarse codes to
prescription-level decision making over detailed clinical
trajectories and specific medication-dose-route triples, \methodName
captures the per-timepoint, information-rich nature of real
prescribing. Our reasoning-chain perturbation method generates
patient-specific distractors that require clinical reasoning rather
than knowledge recall. Experiments on 16 LLMs confirm that
\methodName is highly discriminative and evaluates capabilities
distinct from existing medical benchmarks, while case-study analysis reveals systematic oversight and reasoning errors even in frontier models.

{
\small
\bibliography{references.bib}
\bibliographystyle{abbrvnat}
}

\clearpage
\newpage

\appendix

\section{Related Works}
\label{app:related_works}

\textbf{Medical QA Benchmarks.}
A growing family of benchmarks evaluates LLM medical competence through multiple-choice question answering. MedQA~\citep{jin2021disease} and MedMCQA~\citep{pal2022medmcqa} draw from medical licensing examinations (USMLE and AIIMS/NEET, respectively), while PubMedQA~\citep{jin2019pubmedqa} tests reasoning over biomedical abstracts, MMLU~\citep{hendrycks2020measuring} includes several medical subsets, and RxSafeBench~\citep{zhao2025rxsafebench} evaluates contraindication and interaction awareness in simulated consultations.
These benchmarks share a common design: brief clinical stems, single correct answers, and fixed four- or five-option formats. As a result, they primarily assess factual knowledge recall, and frontier LLMs now exceed 90\% accuracy on MedQA~\citep{singhal2023large}, indicating saturation.
The proposed \methodName differs from all of the above along several dimensions: each question requires reasoning over a richly detailed, time-ordered clinical trajectory (median 3{,}111 tokens, up to 118{,}805); the multi-answer MCQ format requires selecting a variable-size medication subset from 7 options on average; and patient-specific distractors generated via reasoning-chain perturbation ensure that questions cannot be solved by general medical knowledge alone. 

\textbf{Medication Recommendation.}
Early work on medication recommendation formulates the problem as admission-level multi-label prediction over structured EHR codes, where each admission is represented by multi-hot vectors of diagnoses, procedures, and medications~\citep{shang2019gamenet, yang2021safedrug, du2025integrating}. GAMENet~\citep{shang2019gamenet} incorporates drug--drug interaction (DDI) knowledge via graph convolutional networks, while SafeDrug~\citep{yang2021safedrug} further encodes drug molecular structure with an MPNN to improve safety-aware prescribing. SubRec~\citep{du2025integrating} introduces patient subgroup modeling for more personalized recommendations. More recently, LLM-based approaches have begun to exploit clinical notes for richer patient representations. 
CliBench~\citep{ma2024clibench} evaluates LLMs on medication prescription from discharge summaries.
LAMO~\citep{zhao2025finegrain} adapts LLMs via medication-aware fine-tuning to reduce overprescription, LEADER~\citep{liu2024large} distills an LLM with a classification head for efficiency, and FLAME~\citep{fan2025finegrained} proposes a list-wise alignment framework with DDI-aware reward shaping.

Despite methodological progress, all of these methods share the same \emph{evaluation protocol}: they operate on a small, clustered medication set (131--182 medications mapped to ATC-3 therapeutic codes) (\autoref{tab:benchmark_comparison}), adopt admission-level temporal granularity that merges multiple independent prescribing decisions into a single label set.
DTR~\citep{li2025time} and REFINE~\citep{bhoi2023refine} partially address temporal modeling and lab-trend features, respectively, but still use the same evaluation setup.
In contrast, \methodName shifts the evaluation to \emph{prescription-level} granularity with 969 unique medications, evaluating whether models can reason over a patient's evolving clinical trajectory to assemble the correct medication set at each prescribing timepoint.

\section{Models}
\label{app:models}
We evaluate 16 LLMs spanning eight proprietary and eight open-source models.
The proprietary set includes four GPT models (GPT-4o~\citep{hurst2024gpt}, GPT-5-Mini~\citep{singh2025openai}, GPT-5~\citep{singh2025openai}, GPT-5.4~\citep{openai2026gpt54thinking}) and four Gemini models (Gemini-2.5-Flash~\citep{comanici2025gemini25}, Gemini-2.5-Pro~\citep{comanici2025gemini25}, Gemini-3-Flash~\citep{gemini3_google_2025}, Gemini-3.1-Pro~\citep{google2026gemini31pro}).
The open-source set includes general-purpose models of varying scales
(Qwen2.5-72B-Instruct~\citep{yang2024qwen}, Gemma-3-27B-IT~\citep{gemmateam2025gemma3}, Qwen3.6-27B~\citep{yang2025qwen3}, Llama-3.1-8B-Instruct~\citep{grattafiori2024llama}, Qwen2.5-7B-Instruct~\citep{yang2024qwen}) and three medical-domain models fine-tuned from the corresponding base checkpoints of the above families:
HuatuoGPT-o1-72B~\citep{chen2024huatuogpt} from the Qwen2.5 family,
HuatuoGPT-o1-8B~\citep{chen2024huatuogpt} from the Llama-3.1 family,
and MedGemma-27B~\citep{sellergren2025medgemma} from the Gemma-3
family.
Since both the Instruct and medical variants originate from the
same base weights, this pairing isolates the effect of
medical-domain fine-tuning versus general-purpose instruction
tuning.

\section{Case Study: Frontier Model Errors}
\label{app:case_study}

\begin{figure}[t]
  \centering
  \includegraphics[width=1\linewidth]{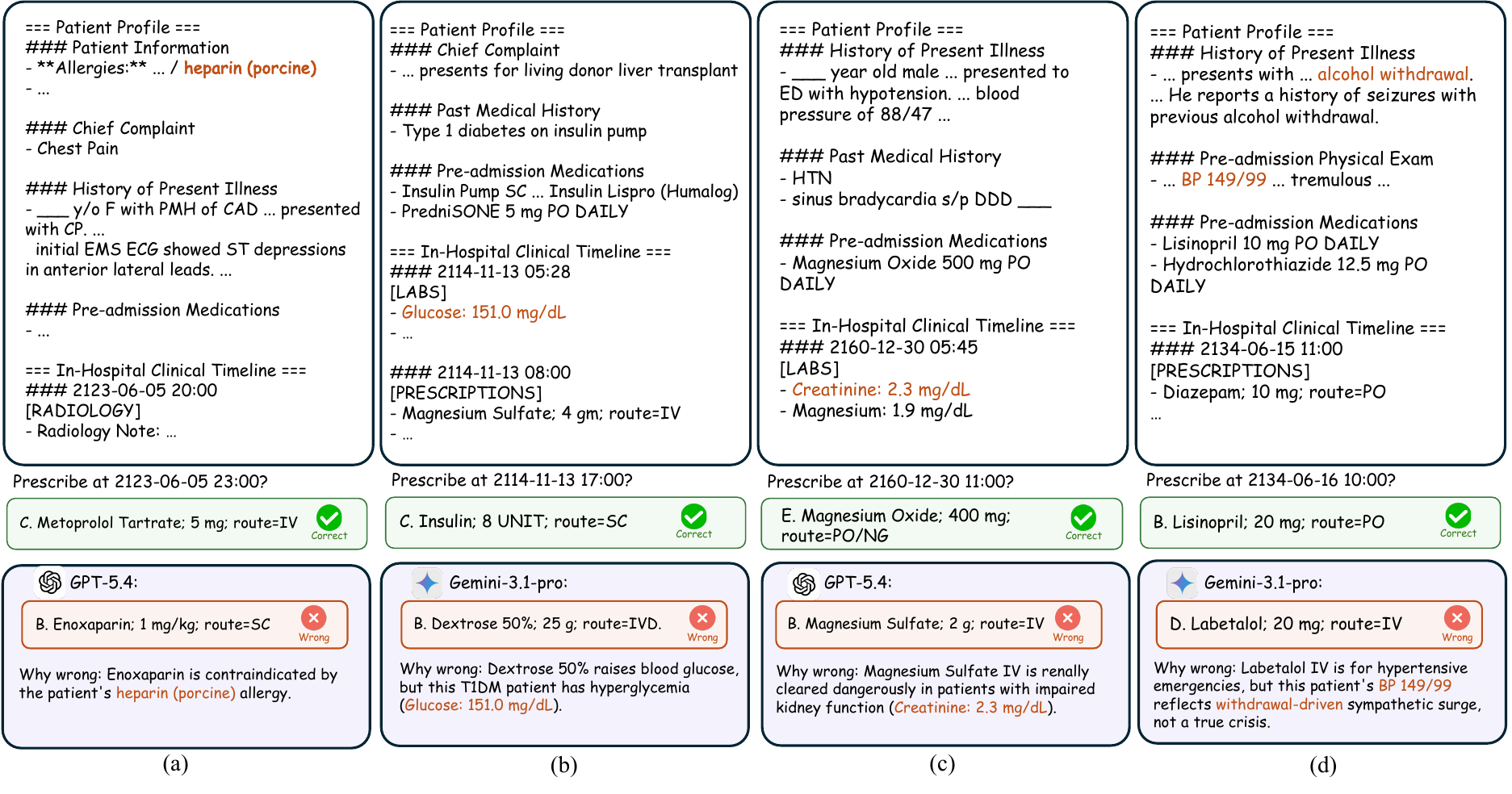}
  \caption{Representative failure cases from GPT-5.4 and Gemini-3.1-Pro on \methodName. (a)--(b) illustrate oversight errors, where the model overlooks explicitly stated patient information. (c)--(d) illustrate reasoning errors, where the model fails to derive a clinical conclusion from raw observations.}
  \label{fig:case_study}
\end{figure}

To illustrate the types of errors that persist in frontier LLMs, we present representative failure cases from GPT-5.4 and Gemini-3.1-Pro in~\autoref{fig:case_study}.
We identify two recurring error patterns: \emph{oversight errors}, where the model ignores explicitly stated patient information (e.g., a documented allergy or a current medication), and \emph{reasoning errors}, where the model fails to derive a clinically necessary conclusion from raw observations (e.g., inferring renal impairment from laboratory values and adjusting the prescription accordingly).

\section{Data Access}
\label{app:data_access}
Individuals interested in accessing \methodName must complete a training course in research with human participants and sign a data use agreement (DUA). The DUA requires users to adequately safeguard the dataset, to not attempt to reidentify individuals, to not share the data, and to report issues relating to deidentification. \footnote{*** add that further details will be provided after the paper is published. // we have provided the data to the reviewer.}

\section{Timepoint Selection}
\label{app:timepoint_selection}
 
Not every prescribing event in raw EHR data reflects an active clinical decision.
Continuations of pre-admission medications and routine standing orders, such as prophylactic agents, vitamins, and vaccines, follow protocol rather than respond to the patient's evolving clinical context, and thus carry little decision value for evaluation.
We therefore apply a two-step filtering procedure to retain only clinically meaningful prescribing timepoints.
 
\textbf{Step 1: Medication-level classification.}
For each candidate timepoint, we prompt an LLM to classify every medication prescribed at that timepoint into one of three categories:
\begin{itemize}[leftmargin=1.2em, itemsep=2pt, topsep=2pt]
\item \textsc{Reasoned\_New}: a medication newly initiated in response to the clinical trajectory $\mathcal{H}_{a,<t}$, reflecting an active prescribing decision.
\item \textsc{Continuation}: a medication that the patient was already taking prior to admission and is simply continued during the hospital stay.
\item \textsc{Routine}: a standing or protocol-driven order (e.g., prophylactic heparin, multivitamins) that is independent of the patient's specific trajectory.
\end{itemize}
A timepoint is retained only if it contains at least one \textsc{Reasoned\_New} medication, ensuring that the prescribing decision at $(a, t)$ is genuinely conditioned on the trajectory $\mathcal{H}_{a,<t}$ rather than reducible to admission-time information alone.
The timepoint prompt is provided in ~\appref{app:prompt_extract_data}.
 
\textbf{Step 2: Temporal balancing.}
For each admission, we partition the qualifying timepoints by their relative positions within the admission into early, middle, and late terciles, and sample from each tercile.
This ensures that the benchmark covers different stages of inpatient care, from initial stabilization decisions to later-phase adjustments based on accumulated clinical evidence.
The final dataset contains 31.4\% early, 32.9\% middle, and 35.7\% late-phase questions (\autoref{tab:basic_stats}).

\section{Additional Results}
\label{app:additional_result}

\subsection{Allergy-Repeated Probing on Oversight-Error Cases}
\label{app:allergy_repeat}

To test whether oversight errors (\autoref{sec:failure_analysis}) reflect a faithfulness limitation rather than a knowledge gap, we conduct a probing experiment on the subset of MCQs where a model selected a medication listed in the patient's allergy field.
For each such case, we re-run the question with the patient's allergy list appended verbatim to the end of the prompt, just before the answer field.
\autoref{tab:allergy_repeat} reports the proportion of cases in which the model no longer selected any allergic option after this intervention.

\begin{table}[h]
\centering
\small
\caption{Allergy-repeated probing on oversight-error cases. For each MCQ in which a model selected a medication the patient was documented to be allergic to, we re-run the question with the allergy list repeated at the end of the prompt. $N$: number of such cases. ``Fixed'': the model no longer selected any allergic option in the rerun.}
\label{tab:allergy_repeat}
\begin{tabular}{lccc}
\toprule
Model & $N$ & Fixed & Fix rate \\
\midrule
Gemini-3-Flash & 70 & 59 & 84.3\% \\
Gemini-2.5-Flash & 67 & 58 & 86.6\% \\
\midrule
\textbf{Total} & \textbf{137} & \textbf{117} & \textbf{85.4\%} \\
\bottomrule
\end{tabular}
\end{table}

Simply repeating the allergy list at the end of the prompt resolves 117 of 137 errors (85.4\%).
This is consistent with the interpretation that oversight errors arise primarily from the model failing to attend to the allergy field located earlier in a long prompt, rather than from missing knowledge about which medications the patient cannot receive.

\subsection{F1 Scores by Temporal Phase}
\autoref{tab:temporal_results} reports the F1 scores
by temporal phase, corresponding to \autoref{fig:temporal_phase}
in the main text.

\begin{table*}[h]
\centering
\small
\caption{
F1 scores at
different temporal phases.}
\label{tab:temporal_results}
\begin{NiceTabular}{cl|ccc}
\toprule
 & Model & F1\textsubscript{early} & F1\textsubscript{mid} & F1\textsubscript{late} \\
\midrule
\Block{8-1}{\rotate \textit{Proprietary}}
 & Gemini-3.1-Pro       & \textbf{77.80} & \textbf{77.40} & \textbf{76.10} \\
 & GPT-5.4              & 74.70 & 70.60 & 70.70 \\
 & GPT-5                & 71.30 & 72.10 & 71.80 \\
 & GPT-5-Mini           & 72.50 & 71.60 & 69.90 \\
 & Gemini-3-Flash       & 75.10 & 68.80 & 68.30 \\
 & Gemini-2.5-Pro       & 72.20 & 70.10 & 68.50 \\
 & Gemini-2.5-Flash     & 69.70 & 68.80 & 64.00 \\
 & GPT-4o               & 60.80 & 58.50 & 57.50 \\
\midrule
\Block{8-1}{\rotate \textit{Open}}
 & Qwen3.6-27B          & \underline{72.10} & \underline{73.60} & \underline{69.60} \\
 & HuatuoGPT-o1-72B     & 59.40 & 58.40 & 54.50 \\
 & MedGemma-27B         & 57.30 & 57.30 & 56.90 \\
 & Qwen2.5-72B-Instruct & 58.20 & 55.90 & 54.30 \\
 & Gemma-3-27B-IT       & 57.00 & 54.40 & 53.90 \\
 & Qwen2.5-7B-Instruct  & 49.80 & 54.80 & 54.10 \\
 & HuatuoGPT-o1-8B      & 46.33 & 47.96 & 45.33 \\
 & Llama-3.1-8B-Instruct & 46.36 & 45.38 & 43.96 \\
\bottomrule
\end{NiceTabular}
\end{table*}

\subsection{Design Validation}
\label{app:design_validation}
 
\footnote{*** either mention this subsec in the main text, or remove it // We want to retain it in case the reviewers ask questions.}
We conduct experiments to validate the input components (patient profile $\mathcal{P}_a$ and clinical trajectory $\mathcal{H}_{a,<t}$) defined in \autoref{sec:task_input}.

\begin{table}[h]
\centering
\small
\setlength{\tabcolsep}{6pt}
\caption{Ablation of patient context on \textsc{\methodName} (GPT-5-Mini).
Best values in \textbf{bold}.}
\label{tab:ablation_context}
\begin{tabular}{l ccccc}
\toprule
& Exact Match & Jaccard & Precision & Recall & F1 \\
\midrule
\methodName       & \textbf{42.7} & \textbf{64.6} & \textbf{76.5} & \textbf{72.3} & \textbf{71.3} \\
\quad w/o $\mathcal{P}_a$  & 33.2 & 55.3 & 69.3 & 64.8 & 62.9 \\
\quad w/o $\mathcal{H}_{a, <t}$ & 33.4 & 56.6 & 68.2 & 66.1 & 63.9 \\
\bottomrule
\end{tabular}
\end{table}
 
\textbf{Both input components provide complementary information.}
\autoref{tab:ablation_context}
shows the effect of removing each input component using GPT-5-Mini.
Removing the patient profile $\mathcal{P}_a$ drops F1 by 8.4 points, and removing the clinical trajectory $\mathcal{H}_{a,<t}$ drops F1 by 7.4 points.
The two ablations degrade performance through different mechanisms: without $\mathcal{P}_a$, the model loses baseline constraints such as allergies and chronic conditions; without $\mathcal{H}_{a,<t}$, it loses the evolving clinical context that motivates the prescription.
This confirms that prescribing decisions in \methodName require integrating both types of clinical information, consistent with the task formulation in \autoref{sec:task_input}.

\section{Limitations}
\label{app:limitations}
\methodName is constructed from MIMIC-IV, a US tertiary hospital
dataset, so certain medications in the benchmark may not be
available in all regions due to differences in medication approval and
formulary practices.

\section{Impact Statement}
\label{app:impact_statement}

\methodName provides a rigorous evaluation tool for assessing LLM prescribing capability against realistic clinical scenarios.
By identifying concrete failure patterns in frontier models, such as overlooking documented allergies and failing to derive clinical conclusions from observations, the benchmark offers actionable guidance for developing safer and more reliable clinical AI systems.
The benchmark is constructed from de-identified data under a data use agreement that prohibits re-identification.
We position RxEval as a valuable reference for assessing LLM capabilities in medication recommendation, but it cannot substitute for the rigorous review and validation processes of medical institutions before any clinical deployment.

\section{Prompts}
\label{app:prompts}

\subsection{Prompt for Extracting MIMIC-IV EHR Data}
\label{app:prompt_extract_data}

\begin{prompt}{Patient-Profile Extraction}{patient_profile_extraction}
TASK \\
Extract ONLY information that is true BEFORE the current hospitalization (i.e., history, baseline status, chronic conditions, prior meds, allergies, prior procedures). \\
DO NOT include hospital-course events such as new inpatient meds, inpatient labs/imaging results, procedures performed during this visit, or discharge instructions. \\

SOURCE NOTE \\
\{Structured Patient Information\} \\
\{Clinical Note\} \\

OUTPUT RULES \\
- Use the exact headings and bullet structure below. \\
- If a field is not present, write: ``Unknown''. \\
- If the note contains placeholders like ``\_\_\_'', keep them as ``Unknown'' (do not guess). \\
- Prefer copying exact phrases; do not paraphrase diagnoses into new terms. \\
- History of Present Illness is important, try to detail all the information in the note. \\
- Physical Exam is conducted before hospitalization, include all its findings. \\
- Pre-visit Medications should be those taken prior to visit only, you can also extract them from ``Brief Hospital Course'' if you think they are taken prior to visit. \\

\#\#\# Patient Information \\
- \textbf{Age:} ... \\
- \textbf{Sex:} ... \\
- \textbf{Service:} ... \\
- \textbf{Allergies:} ... \\
- ... \\

\#\#\# Chief Complaint \\
- ... \\

\#\#\# History of Present Illness \\
- ... \\

\#\#\# Review of Systems (Pre-visit) \\
- ... \\

\#\#\# Past Medical History \\
- ... \\

\#\#\# Social History \\
- ... \\

\#\#\# Family History \\
- ... \\

\#\#\# Pre-visit Physical Exam \\
- ... \\

\#\#\# Pre-visit Medications \\
- ... \\
\end{prompt}

\subsection{Prompts for Constructing Multiple-Choice Questions}
\label{app:construct_mcq}

\begin{prompt}{Timepoint Selection}{timepoint_select_prompt}
\begin{lstlisting}[
  basicstyle=\small\ttfamily,
  breaklines=true,
  xleftmargin=0pt,
  xrightmargin=0pt,
  frame=none,
  aboveskip=0pt,
  belowskip=0pt,
  columns=fullflexible,
  keepspaces=true,
]
TASK: Classify each medication at this timepoint and assess whether it can support a clinical reasoning MCQ.

INPUTS:
[PATIENT PROFILE] {patient_profile}
[IN-HOSPITAL CLINICAL TIMELINE] {clinical_trajectory}
[MEDICATIONS AT {prescription_time}] {actual_prescriptions}

For each medication, classify:

- REASONED_NEW: New drug or significant dose/route change driven by a specific clinical event DOCUMENTED in the hospital timeline (lab, imaging, exam change, clinical deterioration/improvement). Must go beyond "patient was already on this at home."
- CONTINUATION: Drug on the pre-admission medication list being continued into inpatient orders with no new in-hospital clinical trigger.
- ROUTINE: Protocol-driven, auxiliary, or infrastructure medication (e.g., NS flush, standard DVT prophylaxis, paired bowel regimen, vaccine).

For each medication, also provide:
- timeline_dependent: true if justifying this prescription REQUIRES referencing a specific in-hospital event.
- triggering_event: (REASONED_NEW only) the documented event with timestamp that drives this prescription.

DECISION LOGIC:
1. SKIP if one medication name appears twice.
2. SKIP if no medication is REASONED_NEW with
   timeline_dependent=true.
3. PROCEED otherwise.

OUTPUT (JSON):
{
  "prescription_time": "...",
  "medications": [
    {
      "medication": "<drug>",
      "classification": "REASONED_NEW|CONTINUATION|ROUTINE",
      "timeline_dependent": true|false,
      "triggering_event": "<timestamp + event, or null>",
      "reason": "<1 sentence>"
    }
  ],
  "decision": "PROCEED|SKIP",
  "reason": "<1 sentence>"
}
\end{lstlisting}
\end{prompt}

\begin{prompt}{Reasoning Chain Annotation}{reasoning_chain_annotation}
\begin{lstlisting}[
  basicstyle=\small\ttfamily,
  breaklines=true,
  xleftmargin=0pt,
  xrightmargin=0pt,
  frame=none,
  aboveskip=0pt,
  belowskip=0pt,
  columns=fullflexible,
  keepspaces=true,
]
TASK: For each medication ordered at a timepoint, trace the clinical reasoning from patient data to prescribing decision using an evidence chain.

INPUTS:
[PATIENT PROFILE] {patient_profile}
[IN-HOSPITAL CLINICAL TIMELINE] {clinical_trajectory}
[MEDICATIONS AT {prescription_time}] {actual_prescriptions}

RULES:
- Never cite a medication's own order as evidence for itself.
- If a medication is auxiliary to another at the same timepoint (e.g., PPI with corticosteroids), reference that medication.
- Facts must be verifiable against the inputs. Do not fabricate.
- If no supporting evidence exists in the timeline, return an empty chain_steps array.

OUTPUT (JSON):
{
  "prescription_time": "{prescription_time}",
  "reasoning_chains": [
    {
      "medication": "<drug>",
      "dose": "<dose>",
      "route": "<route>",
      "chain_steps": [
        {
          "step": <int>,
          "fact": "<from input data>",
          "source": "<e.g., 'timeline: 2024-01-03 06:00
                      lab results'>",
          "depends_on": [],
          "inference": "<what this implies, combined with prior steps>"
        }
      ],
      "summary": "<2-3 sentence overall summary of why this medication was prescribed, integrating all chain steps into a coherent clinical rationale>"
    }
  ]
}
\end{lstlisting}
\end{prompt}

\begin{prompt}{Reasoning Chain Critic}{reasoning_chain_critic}
\begin{lstlisting}[
  basicstyle=\small\ttfamily,
  breaklines=true,
  xleftmargin=0pt,
  xrightmargin=0pt,
  frame=none,
  aboveskip=0pt,
  belowskip=0pt,
  columns=fullflexible,
  keepspaces=true,
]
TASK: Verify that each reasoning chain is factually grounded in the provided data and logically coherent.

INPUTS:
[PATIENT PROFILE] {patient_profile}
[IN-HOSPITAL CLINICAL TIMELINE] {clinical_trajectory}
[MEDICATIONS AT {prescription_time}] {actual_prescriptions}
[REASONING CHAINS] {reasoning_chains}

CHECK EACH CHAIN FOR:
1. Every cited fact exists in the inputs with correct values.
2. No step cites its own medication or events after {prescription_time}.
3. Each non-root step genuinely builds on its depends_on steps, not just introduces an independent fact.
4. Inferences follow from the cited facts: no unsupported leaps, and no treating mere temporal association as causal rationale.
5. Each inference must be medically correct given the cited facts (right interpretation of labs/conditions; drug is genuinely indicated).
6. The chain's cumulative evidence must be SUFFICIENT to justify prescribing this specific medication. REJECT if a chain with no independent triggering event (empty or single-step restatement of the prescription).
7. The chain's summary and all inference fields must make a committed assertion about the core indication: not concede it is unknown or speculative (e.g., "unclear", "presumably", "may have been given for"). Uncertainty about peripheral details is acceptable; REJECT if uncertainty about the core indication.

CHECK THE ENTIRE TIMEPOINT:
1. At least one chain must be driven by a new in-hospital event from the timeline, not just admission info, home meds, or standard protocols. If none is, REJECT the entire timepoint.
2. No two chains may contradict each other on the same clinical fact or its interpretation.

KEY PRINCIPLE: Clinically plausible does not equals to evidenced. If reasoning is sensible but not supported by THIS patient's data, REVISE it.
A chain that openly admits uncertainty about why the drug was prescribed is NOT a valid chain, regardless of how well the surrounding text is written, REJECT it.

OUTPUT (JSON):
{
  "prescription_time": "{prescription_time}",
  "per_medication": [
    {
      "medication": "<drug>",
      "verdict": "PASS|REVISE|REJECT",
      "issues": [
        {
          "step": <step number or "summary">,
          "type": "fabricated_fact|wrong_value|temporal_violation|self_justification|broken_dependency|logical_leap|insufficient_evidence|self_negation|empty_chain",
          "detail": "<what is wrong; for self_negation, quote the exact phrase>"
        }
      ],
      "feedback": "<specific instructions for fixing the chain,
                    only if REVISE>"
    }
  ],
  "overall_verdict": "PASS|REVISE|REJECT"
}

VERDICTS:
- PASS: All facts verified, logic valid, rationale committed (no self-negation).
- REVISE: Fixable issues: feed the feedback back to the chain builder for retry.
- REJECT: Unfixable: fabricated core facts, fundamentally invalid reasoning, empty chain without trigger event, OR the summary/inferences self-negate the rationale. No retry.
\end{lstlisting}
\end{prompt}

\begin{prompt}{Distractor Generation}{distractor_generation}
\begin{lstlisting}[
  basicstyle=\small\ttfamily,
  breaklines=true,
  xleftmargin=0pt,
  xrightmargin=0pt,
  frame=none,
  aboveskip=0pt,
  belowskip=0pt,
  columns=fullflexible,
  keepspaces=true,
]
TASK: Given correct medications with validated evidence chains, generate {num_distractors} plausible but clinically INCORRECT medication options.

INPUTS:
[PATIENT PROFILE] {patient_profile}
[IN-HOSPITAL CLINICAL TIMELINE] {clinical_trajectory}
[CORRECT MEDICATIONS WITH EVIDENCE CHAINS] {formatted_correct_meds}
{feedback_section}

REQUIREMENTS:
- Each distractor must be a real medication with a realistic dose and route.
- Each must target a specific vulnerability in a correct medication's evidence chain: a step where misreading the evidence would make the distractor seem correct.
- Each must be **clearly wrong** for THIS patient at THIS timepoint, not a genuinely acceptable alternative.
- No duplicates, no correct medications reused as distractors.
- The "medication" field must contain ONLY the drug name (e.g., "Metoprolol"). Put dose and route in their own fields.

DESIGN STRATEGIES (non-exhaustive, mix as appropriate):
- Same drug class but wrong specific agent for this patient
- Right symptom but wrong mechanism or drug class
- Contraindicated given this patient's specific profile
- Appropriate at a different treatment phase but not now
- Addresses a real condition but not indicated at this timepoint

OUTPUT (JSON):
{
  "distractors": [
    {
      "targets_correct_medication": "<which correct med>",
      "targets_evidence_step": <step number or null>,
      "why_plausible": "<1 sentence>",
      "why_wrong": "<1 sentence>",
      "medication": "<drug>",
      "dose": "<dose>",
      "route": "<route>"
    }
  ]
}
\end{lstlisting}
\end{prompt}

\begin{prompt}{Distractor Validation}{distractor_validation}
\begin{lstlisting}[
  basicstyle=\small\ttfamily,
  breaklines=true,
  xleftmargin=0pt,
  xrightmargin=0pt,
  frame=none,
  aboveskip=0pt,
  belowskip=0pt,
  columns=fullflexible,
  keepspaces=true,
]
TASK: Verify each distractor's rejection reason is correct with the patient's clinical state.

INPUTS:
[PATIENT PROFILE] {patient_profile}
[IN-HOSPITAL CLINICAL TIMELINE] {clinical_trajectory}
[DISTRACTORS] {distractor_meds_detailed}

For each distractor, perform these steps in order.

STEP 1 - Fact & mechanism grounding
(i) If why_wrong cites any patient-specific fact, it MUST appear in the inputs. If not found: INVALID. Institutional claims ("formulary preference", "hospital protocol", "not stocked") are never acceptable: also INVALID.
(ii) If why_wrong cites a pharmacological mechanism, verify it is a well-established property of THIS distractor drug, not mis-attributed from another drug. If wrong: INVALID.

STEP 2 - Difficulty check
If the distractor can be rejected without any patient-specific data (wrong drug class entirely, mechanism opposite to therapeutic goal, trivially non-therapeutic in context): TOO_EASY.

If none of the above triggers: GOOD.

OUTPUT (JSON):
{
  "distractor_evaluations": [
    {
      "medication": "<drug>",
      "fact_grounding_check": "<what fact cited? found in record? quote supporting line or 'NOT_FOUND'>",
      "mechanism_accuracy_check": "<what mechanism claimed? accurate for this drug? 'OK' or correction>",
      "distractor_claim": "<extracted rejection mechanism/fact>",
      "targets_correct_medication": "<which correct med>",
      "quality": "GOOD|TOO_EASY|INVALID",
      "reason": "<1 sentence to explain the reason>",
      "suggestion": "<replacement direction if INVALID/TOO_EASY, else null>"
    }
  ],
  "overall_verdict": "PASS|FAIL",
  "feedback": "<regeneration guidance if FAIL, else null>"
}

PASS if all distractors are GOOD. FAIL otherwise.
\end{lstlisting}
\end{prompt}

\begin{prompt}{MCQ Question}{mcq_prompt}
\begin{lstlisting}[
  basicstyle=\small\ttfamily,
  breaklines=true,
  xleftmargin=0pt,
  xrightmargin=0pt,
  frame=none,
  aboveskip=0pt,
  belowskip=0pt,
  columns=fullflexible,
  keepspaces=true,
]
A patient presents with the following profile and clinical timeline.
Based on the clinical data available, which of the following medications represent appropriate therapeutic decisions at timepoint {prescription_time}? (Select ALL that apply)

=== Patient Profile ===
{patient_profile}

=== In-Hospital Clinical Timeline ===
{clinical_trajectory}

Options:
{options_text}

Based on the clinical evidence provided, select ALL correct options.
Respond with ONLY the option letter(s), comma-separated.
Example: A, C, E
\end{lstlisting}
\end{prompt}

\subsection{Prompt for Constructing Open-ended Questions}
\label{app:prompt_openended_query}
\begin{prompt}{Open-ended Questions}{open_ended_query}
\begin{lstlisting}[
  basicstyle=\small\ttfamily,
  breaklines=true,
  xleftmargin=0pt,
  xrightmargin=0pt,
  frame=none,
  aboveskip=0pt,
  belowskip=0pt,
  columns=fullflexible,
  keepspaces=true,
]
Given a patient's profile and clinical timeline, recommend the medications that should be prescribed at the target decision point.

=== Patient Profile ===
{patient_profile}

=== In-Hospital Clinical Timeline ===
{clinical_trajectory}

Based on the patient profile and clinical timeline above, what medication(s) should be ordered at {prescription_time}?

For each medication, provide on a new line in the following format:
  - <drug name>; <dose>; route=<route>

Example:
  - Heparin; 5000 UNIT; route=SC
  - Acetaminophen; 650 mg; route=PO

\end{lstlisting}
\end{prompt}





\section{RxEval Example}
\label{app:examples}
Due to the data use agreement of MIMIC-IV, we cannot display actual patient records. The following shows 
a synthetic example that mirrors the structure and clinical patterns observed in real cases while containing no real patient data.

\begin{exmp}{}{example_rxeval}
\begin{lstlisting}[
  basicstyle=\small\ttfamily,
  breaklines=true,
  xleftmargin=0pt,
  xrightmargin=0pt,
  frame=none,
  aboveskip=0pt,
  belowskip=0pt,
  columns=fullflexible,
  keepspaces=true,
  escapeinside={(*@}{@*)},
]
A patient presents with the following profile and clinical timeline.
Based on the clinical data available, which of the following medications represent appropriate therapeutic decisions at timepoint 2145-03-17 14:00? (Select ALL that apply)
 
=== Patient Profile ===
### Patient Information
- **Age:** 68
- **Sex:** M
- **Service:** MEDICINE
- **Allergies:** Penicillins (rash)
- **Race:** BLACK/AFRICAN AMERICAN
- **Marital Status:** MARRIED
- **Language:** English
- **Insurance:** Medicare
 
### Chief Complaint
- Fever, cough, shortness of breath
 
### History of Present Illness
- Mr. ___ is a ___ year old male with a history of COPD, type 2 diabetes, CKD stage III, gout, and atrial fibrillation on warfarin who presents with three days of progressive cough productive of yellow-green sputum, fevers to 102(*@$^\circ$@*)F at home, and worsening dyspnea. He reports that his symptoms began with a sore throat and rhinorrhea, followed by the onset of chills and rigors. He notes increased dyspnea on exertion and is now short of breath at rest. He denies chest pain, hemoptysis, nausea, vomiting, or diarrhea. He reports compliance with all medications. His last INR check was approximately two weeks ago and was in the therapeutic range.
 
### Review of Systems (Pre-admission)
- Denies chest pain, palpitations, or syncope.
- Denies abdominal pain, nausea, vomiting, or diarrhea.
- Denies dysuria or hematuria.
- Denies joint pain or swelling.
- Reports decreased appetite over the past two days.
- Reports subjective fevers and chills at home.
 
### Past Medical History
- COPD (moderate, on home inhalers, no home O2)
- Type 2 Diabetes Mellitus (on oral agents, HbA1c 7.8% three months ago)
- Chronic Kidney Disease stage III (baseline Cr 1.6-1.8)
- Gout (on prophylactic allopurinol, last flare > 1 year ago)
- Atrial fibrillation (on warfarin, rate-controlled)
- Hypertension
- Gastroesophageal reflux disease
- Obesity (BMI 34)
- s/p appendectomy ___
 
### Social History
- Unknown
 
### Family History
- Father died of stroke at age ___. Mother had type 2 diabetes and died of complications of heart failure. Brother with COPD.
 
### Pre-admission Physical Exam
- **VS:** T 101.8(*@$^\circ$@*)F, BP 98/62, HR 108 (irregularly irregular), RR 24, SpO2 91% on RA
- **GENERAL:** Ill-appearing male, diaphoretic, speaking in short sentences
- **HEENT:** NCAT, MMM, no JVD
- **NECK:** Supple, no lymphadenopathy
- **CARDIAC:** Irregularly irregular, tachycardic, no murmurs, rubs, or gallops
- **LUNGS:** Decreased breath sounds at right base with dullness to percussion, scattered rhonchi bilaterally, egophony at right lower lung field
- **ABDOMEN:** Soft, NT, ND, +BS, well-healed RLQ scar
- **EXTREMITIES:** Warm, no cyanosis, trace bilateral lower extremity edema
- **NEURO:** Alert and oriented x3, no focal deficits
 
### Pre-admission Medications
- Warfarin 5 mg PO daily
- Metformin 1000 mg PO twice daily
- Linagliptin 5 mg PO daily
- Amlodipine 5 mg PO daily
- Allopurinol 300 mg PO daily
- Albuterol Sulfate HFA 90 mcg 2 PUFF INH every 6 hours as needed
- Tiotropium Bromide 18 mcg INH daily
- Pantoprazole 40 mg PO daily
```
 
=== In-Hospital Clinical Timeline ===
### 2145-03-15 00:00
[PROCEDURES]
- Venous catheterization, not elsewhere classified
 
### 2145-03-15 08:00
[PRESCRIPTIONS]
- CefTRIAXone; 1 g; route=IV
- Azithromycin; 500 mg; route=IV
- 0.9% Sodium Chloride; 1000 mL; route=IV
- Pantoprazole; 40 mg; route=IV
- Tiotropium Bromide; 18 mcg; route=INH
- Albuterol 0.083% Neb Soln; 1 NEB; route=IH
- Insulin; 0 UNIT; route=SC
- Heparin; 5000 UNIT; route=SC
 
### 2145-03-15 08:00
[ENDED PRESCRIPTIONS]
- Warfarin; 5 mg; route=PO
- Metformin; 1000 mg; route=PO
 
### 2145-03-15 08:45
[LABS]
- White Blood Cells [Blood] (Hematology): 18.6 K/uL
- Red Blood Cells [Blood] (Hematology): 4.12 m/uL
- Hemoglobin [Blood] (Hematology): 12.1 g/dL
- Hematocrit [Blood] (Hematology): 36.8 %
- MCV [Blood] (Hematology): 89 fL
- MCH [Blood] (Hematology): 29.4 pg
- MCHC [Blood] (Hematology): 32.9 %
- RDW [Blood] (Hematology): 14.2 %
- Platelet Count [Blood] (Hematology): 245 K/uL
- Neutrophils [Blood] (Hematology): 88 %
- Bands [Blood] (Hematology): 8 %
- Lymphocytes [Blood] (Hematology): 3 %
- Monocytes [Blood] (Hematology): 1 %
- Sodium [Blood] (Chemistry): 139 mEq/L
- Potassium [Blood] (Chemistry): 5.1 mEq/L
- Chloride [Blood] (Chemistry): 102 mEq/L
- Bicarbonate [Blood] (Chemistry): 20 mEq/L
- Anion Gap [Blood] (Chemistry): 22 mEq/L
- Urea Nitrogen [Blood] (Chemistry): 38 mg/dL
- Creatinine [Blood] (Chemistry): 2.3 mg/dL
- Glucose [Blood] (Chemistry): 224 mg/dL
- Calcium, Total [Blood] (Chemistry): 8.6 mg/dL
- Magnesium [Blood] (Chemistry): 2.0 mg/dL
- Phosphate [Blood] (Chemistry): 4.5 mg/dL
- Lactate [Blood] (Chemistry): 3.2 mmol/L
- Procalcitonin [Blood] (Chemistry): 4.8 ng/mL
- PT [Blood] (Hematology): 22.1 sec
- INR(PT) [Blood] (Hematology): 2.8
- PTT [Blood] (Hematology): 38.5 sec
- Albumin [Blood] (Chemistry): 2.8 g/dL
 
### 2145-03-15 09:22
[RADIOLOGY]
- Radiology Note: FINDINGS:
  The heart size is at the upper limits of normal. There is a dense
  consolidation in the right lower lobe with air bronchograms. A small
  right-sided pleural effusion is present. The left lung is clear. No
  pneumothorax is identified. The central venous catheter tip is at the
  cavoatrial junction in satisfactory position.
  
  IMPRESSION:
  Right lower lobe consolidation consistent with pneumonia. Small right
  pleural effusion. Central line in satisfactory position.
 
### 2145-03-15 14:00
[PRESCRIPTIONS]
- 0.9% Sodium Chloride; 1000 mL; route=IV
- Linagliptin; 5 mg; route=PO
 
### 2145-03-15 20:00
[LABS]
- Lactate [Blood] (Chemistry): 2.4 mmol/L
- Creatinine [Blood] (Chemistry): 2.1 mg/dL
- Potassium [Blood] (Chemistry): 4.8 mEq/L
- Glucose [Blood] (Chemistry): 186 mg/dL
 
### 2145-03-16 06:30
[LABS]
- White Blood Cells [Blood] (Hematology): 15.2 K/uL
- Neutrophils [Blood] (Hematology): 82 %
- Bands [Blood] (Hematology): 4 %
- Red Blood Cells [Blood] (Hematology): 3.98 m/uL
- Hemoglobin [Blood] (Hematology): 11.8 g/dL
- Hematocrit [Blood] (Hematology): 35.6 %
- Platelet Count [Blood] (Hematology): 221 K/uL
- Sodium [Blood] (Chemistry): 140 mEq/L
- Potassium [Blood] (Chemistry): 4.5 mEq/L
- Chloride [Blood] (Chemistry): 104 mEq/L
- Bicarbonate [Blood] (Chemistry): 23 mEq/L
- Anion Gap [Blood] (Chemistry): 17 mEq/L
- Urea Nitrogen [Blood] (Chemistry): 34 mg/dL
- Creatinine [Blood] (Chemistry): 2.0 mg/dL
- Glucose [Blood] (Chemistry): 168 mg/dL
- Lactate [Blood] (Chemistry): 1.5 mmol/L
- PT [Blood] (Hematology): 19.8 sec
- INR(PT) [Blood] (Hematology): 2.5
- Calcium, Total [Blood] (Chemistry): 8.4 mg/dL
- Magnesium [Blood] (Chemistry): 1.8 mg/dL
 
### 2145-03-16 10:00
[PRESCRIPTIONS]
- Magnesium Sulfate; 2 gm; route=IV
- Azithromycin; 500 mg; route=PO
 
### 2145-03-16 10:00
[ENDED PRESCRIPTIONS]
- Azithromycin; 500 mg; route=IV
- 0.9% Sodium Chloride; 1000 mL; route=IV
 
### 2145-03-16 18:00
[PRESCRIPTIONS]
- Acetaminophen; 650 mg; route=PO
 
### 2145-03-17 06:15
[LABS]
- White Blood Cells [Blood] (Hematology): 11.8 K/uL
- Neutrophils [Blood] (Hematology): 74 %
- Red Blood Cells [Blood] (Hematology): 4.05 m/uL
- Hemoglobin [Blood] (Hematology): 12.0 g/dL
- Hematocrit [Blood] (Hematology): 36.2 %
- Platelet Count [Blood] (Hematology): 238 K/uL
- Sodium [Blood] (Chemistry): 141 mEq/L
- Potassium [Blood] (Chemistry): 4.2 mEq/L
- Chloride [Blood] (Chemistry): 105 mEq/L
- Bicarbonate [Blood] (Chemistry): 25 mEq/L
- Anion Gap [Blood] (Chemistry): 15 mEq/L
- Urea Nitrogen [Blood] (Chemistry): 30 mg/dL
- Creatinine [Blood] (Chemistry): 1.9 mg/dL
- Glucose [Blood] (Chemistry): 152 mg/dL
- PT [Blood] (Hematology): 18.2 sec
- INR(PT) [Blood] (Hematology): 2.3
- Calcium, Total [Blood] (Chemistry): 8.7 mg/dL
- Magnesium [Blood] (Chemistry): 2.1 mg/dL
- Procalcitonin [Blood] (Chemistry): 1.2 ng/mL
 
### 2145-03-17 09:48
[RADIOLOGY]
- Radiology Note: FINDINGS:
  The right lower lobe consolidation persists but is slightly improved
  compared to prior. The small right pleural effusion is stable. No new
  infiltrates are identified. The left lung remains clear. Central venous
  catheter remains in satisfactory position.
  
  IMPRESSION:
  Slight interval improvement in right lower lobe pneumonia. Stable small
  right pleural effusion.
 
### 2145-03-17 10:00
[ENDED PRESCRIPTIONS]
- CefTRIAXone; 1 g; route=IV
- Pantoprazole; 40 mg; route=IV
 
### 2145-03-17 10:00
[PRESCRIPTIONS]
- Pantoprazole; 40 mg; route=PO

Options:
A. Levofloxacin; 750 mg; route=PO
B. Amoxicillin-Clavulanate; 875 mg; route=PO
C. Enoxaparin Sodium; 40 mg; route=SC
D. Metformin; 1000 mg; route=PO
E. Allopurinol; 300 mg; route=PO
F. Gentamicin; 360 mg; route=IV
G. Amlodipine; 5 mg; route=PO
H. Warfarin; 5 mg; route=PO
I. Metoprolol Tartrate; 25 mg; route=PO

Based on the clinical evidence provided, select ALL correct options.
Respond with ONLY the option letter(s), comma-separated.
Example: A, C, E

---------------------------------------------------------------------------
ANSWER: A, C, E, G
---------------------------------------------------------------------------

\end{lstlisting}
\end{exmp}


\end{document}